%% file: main.tex
\documentclass[10pt,twocolumn,letterpaper]{article}

\usepackage[pagenumbers]{cvpr}

\input{preamble}

\definecolor{cvprblue}{rgb}{0.21,0.49,0.74}
\usepackage[pagebackref,breaklinks,colorlinks,allcolors=cvprblue]{hyperref}
\hypersetup{pdftitle={XShot: Cross-Shot Video Object Insertion from a Reference Video},pdfauthor={Bangxun Tang, Chen Chen}}

\def\confName{CVPR}
\def\confYear{2027}

\title{\ours: Cross-Shot Video Object Insertion from a Reference Video}

\author{Bangxun Tang\\
University of California, Irvine\\
{\tt\small bangxunt@uci.edu}
\and
Chen Chen\thanks{Corresponding author.}\\
University of Central Florida\\
{\tt\small chen.chen@ucf.edu}
}

\begin{document}
\maketitle
\input{fig/teaser}
\input{sec/0_abstract}
\input{sec/1_intro}
\input{sec/2_related}
\input{sec/3_method}
\input{sec/4_experiments}
\input{sec/5_conclusion}
\clearpage
\input{sec/6_llm_statement}
{
    \small
    \bibliographystyle{ieeenat_fullname}
    \bibliography{main}
}

\input{sec/X_suppl}

\end{document}

%% file: preamble.tex
\usepackage{cuted}

\usepackage{currfile}

\usepackage{caption}
\usepackage{animate}
\usepackage{xspace}
\newcommand{\ours}{XShot\xspace}

\usepackage{multirow}
\usepackage{bigdelim}
\usepackage{tabularx}
\usepackage{tikz}
\usepackage{pifont}
\usepackage{colortbl}
\newcommand{\cmark}{\textcolor{green!55!black}{\ding{51}}}
\newcommand{\xmark}{\textcolor{red!70!black}{\ding{55}}}
\definecolor{oursrow}{HTML}{E3EEFA}
\definecolor{headgray}{HTML}{EDEDED}

\renewcommand{\paragraph}[1]{\vspace{.5em}\noindent\textbf{#1}}

%% file: fig/teaser.tex
\begin{strip}
\centering
\begin{animateinline}[poster=first,width=\textwidth]{100}
\includegraphics[width=\textwidth]{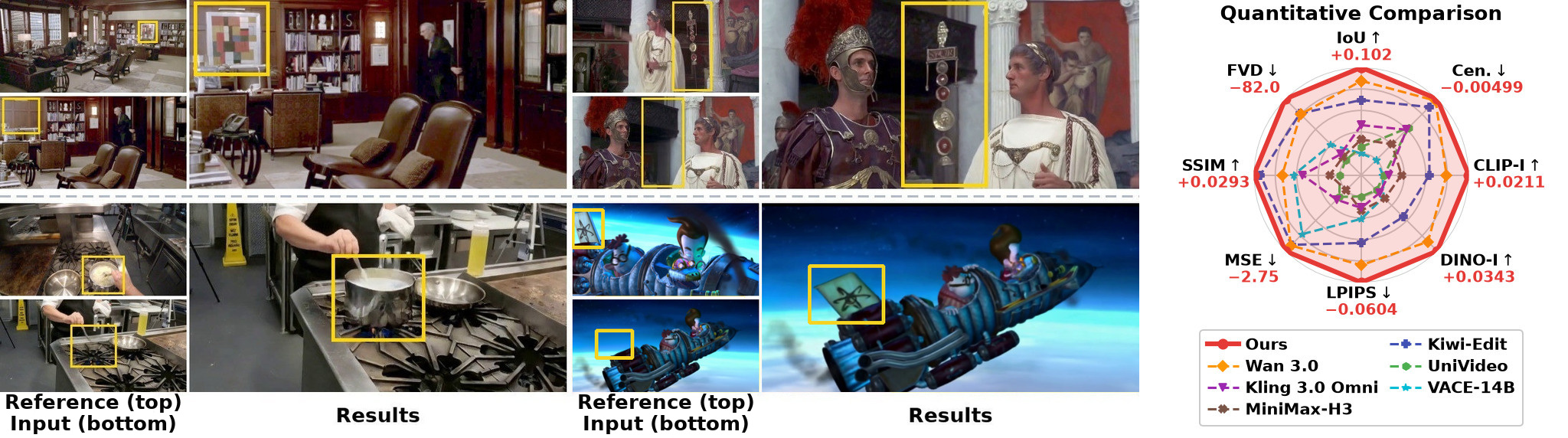}
\newframe[2]
\multiframe{8}{i=0+1}{\includegraphics[width=\textwidth]{figures/teaser/f_\i.jpg}}
\end{animateinline}
\captionsetup{hypcap=false}
\captionof{figure}{Given a reference video from one shot of a scene, a few exemplars of an object and an input video of \textbf{another shot} from which the object is absent, \ours inserts the object at the corresponding location, with the viewpoint, scale and occlusion of the target camera. Yellow boxes mark the object in the reference video, the input video and the result. Readers can click and play the video clips in this figure using \textcolor{red}{\textbf{Adobe Acrobat}}.}
\captionsetup{hypcap=true}
\label{fig:teaser}
\end{strip}

%% file: sec/0_abstract.tex
\begin{abstract}
We present \ours, a framework for cross-shot video object insertion, which adds objects to a shot by watching another shot of the same scene.
Video productions show a scene through many shots, and objects added to one shot must appear consistently in the others.
Most video editors take the appearance of new objects from reference images and their location from a text prompt, a mask or an edited keyframe, leaving this consistency to the user.
\ours reads both from a reference video of another shot, which need not be aligned with the target, and needs no mask, box or camera pose.
We build on a video Diffusion Transformer and formulate the task as in-context generation over the reference video, a few images of the objects and the target video.
A geometry-aware cross-shot attention estimates the epipolar geometry between the two shots from the model's own correspondences and uses it to guide where the target reads the reference and the exemplars.
We train \ours with an erase-and-restore data engine: objects are erased from one shot of a film or a multi-camera recording and restored from another shot of the scene.
Experiments show that \ours places objects more accurately and renders them more faithfully than open-source and commercial video editors, while leaving the rest of the shot intact.
\end{abstract}

%% file: sec/1_intro.tex
\section{Introduction}
\label{sec:intro}

Video diffusion models~\cite{wan,hunyuanvideo,cogvideox} have made generative video editing practical.
Reference-guided editors such as VACE~\cite{vace}, Kiwi-Edit~\cite{kiwiedit} and UniVideo~\cite{univideo} can add an object to a video from a reference image of it.
These editors work on one shot at a time.
A film, however, shows a scene through several shots, each taken from a different camera position.
An object added to one shot must also appear in the other shots, at the same place in the scene.

\begin{figure}[t]
\centering
\includegraphics[width=\columnwidth]{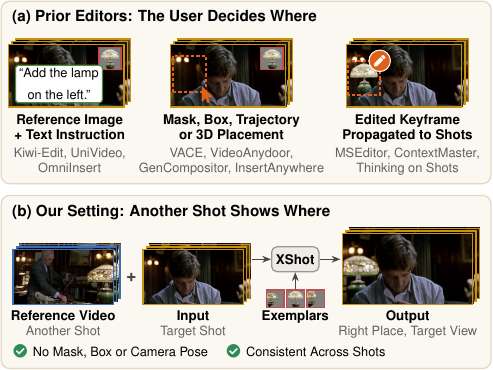}
\caption{Prior editors versus our setting. (a) Prior editors need the user to tell where the object goes in every shot. (b) \ours reads the location from another shot of the scene, so the object stays consistent across shots, with no mask, box or camera pose.}
\label{fig:setting}
\end{figure}

Current tools leave this consistency to the user (\cref{fig:setting}(a)).
Reference-guided editors learn what the object looks like from reference images, but take its location from a text instruction~\cite{kiwiedit,univideo,omniinsert}, or from a mask, a box, a trajectory or a 3D placement given by the user~\cite{vace,videoanydoor,gencompositor,insertanywhere}.
Multi-shot editors~\cite{mseditor,thinkingonshots,contextmaster} propagate an edit across shots, but the edit itself still comes from an instruction or an edited keyframe.
In both cases, the user has to decide where the object goes.
Yet another shot of the scene often already shows the object and its place in the scene.
In a film, for example, a vase may stand on the table in the wide shot of a dialogue but be missing from a close-up of the same scene, a common continuity error.
We therefore introduce a new task called \textbf{cross-shot video object insertion} (\cref{fig:setting}(b)), in which the model reads the location of the object from a reference video, without any mask, box or camera pose.
The location comes from footage of the scene, not from the user.

The two shots need not be taken at the same time.
Between two shots of a film, a static object such as the vase keeps its place, so the reference shows where it stands in the scene.
With synchronized cameras, the reference video also shows where an object is at each moment.
A moving object, such as a ball in a game, can then be inserted as well.

To handle both cases with one model, we present \ours, a video generation model built on a Diffusion Transformer (DiT)~\cite{dit,wan}.
It takes the input video, which is the target shot without the object, the reference video and a few exemplar images of the object, and generates the target shot with the object (\mbox{\cref{fig:teaser}}).
The exemplars show the object as the target camera sees it, because the reference often shows it from another side, far away or partly hidden.
The reference video and the exemplars are placed in the same token sequence as the noisy target video, so every target token can attend to them, and the input video is added to the target tokens~\cite{kiwiedit}.

Reading the location, however, is the main difficulty.
The two shots are taken from different viewpoints, so the object appears at a different image position and size in each.
Attention alone can then pick a region of the reference that only looks similar.
In our experiments, existing editors often misplace the object, give it the wrong size, or copy the view of the reference into the target.
Our key idea is to guide the attention with the geometry of the two cameras.

For two cameras viewing the same scene, epipolar geometry states that a point in one image can only appear along a certain line in the other image, its epipolar line.
In multi-view methods, this line is computed from known camera poses~\cite{epipolartransformers,epidiff,camco,cvd}, which films do not provide.
\ours instead finds matching points between the two videos with a correspondence head that reads them from the attention of the DiT.
From these points, it computes the epipolar lines, and its geometry-aware cross-shot attention makes each target token attend to its epipolar line in the reference.
Target tokens near the line on which the object must appear also attend more to the exemplars.
Since the epipolar geometry already provides much of the spatial relation between the two shots, we pool the reference into $2\times2$ cells, which makes \ours faster to train and to run than the same backbone with the full reference.

Training this model needs a shot with and without the object, together with another shot of the same scene, and no such data exist.
Our erase-and-restore data engine builds them from films and Ego-Exo4D~\cite{egoexo4d}: it selects an object that two shots of a scene both show, erases it from one shot to obtain the input video, and keeps the original as the ground truth and the other shot as the reference.

Our contributions are summarized as follows:
\begin{itemize}
    \item We introduce a new task, cross-shot video object insertion, and present \ours, which reads the location of an object from a reference video and takes its appearance from exemplars, with no mask, box or camera pose.
    \item We propose a geometry-aware cross-shot attention for video Diffusion Transformers. Without camera poses, it estimates the epipolar geometry between two videos from matching points that the model finds itself, and makes the target attend along the epipolar lines, so that objects land at the right place and with the right size.
    \item We build an erase-and-restore data engine that turns films and multi-camera recordings into training data, and a benchmark on which \ours outperforms open-source and commercial editors on every metric, by a large margin in placement and whole-video quality (\mbox{\cref{fig:teaser}}).
\end{itemize}

%% file: sec/2_related.tex
\section{Related Work}
\label{sec:related}

\paragraph{Video diffusion models and in-context conditioning.}
Video generation has moved from image diffusion models with temporal layers~\cite{videoldm,svd} to Diffusion Transformers~\cite{dit} over spatio-temporal patches~\cite{cogvideox,hunyuanvideo,wan}.
A Diffusion Transformer can take conditions in context, by appending their tokens to the sequence.
In-context conditioning was first explored for images~\cite{incontextlora,ominicontrol,easycontrol,uno}.
It has since been adopted for video editing and control~\cite{unic,fulldit,vap}, including background and character replacement in film shots~\cite{paistudio,paiactor}, and for multi-shot generation~\cite{lct,holocine,pecontext}.
Several methods also adapt the position encoding of the conditions.
OminiControl~\cite{ominicontrol} and GenCompositor~\cite{gencompositor} offset the positions of a condition that is not spatially aligned with the target.
Video-As-Prompt~\cite{vap} shifts a reference video in time, PoCo~\cite{pecontext} routes tokens between references and shots, and PAI-Studio~\cite{paistudio} gives a reference image the temporal position of its target frame.
In contrast, our reference shows the same scene from another camera.
We therefore place it beside the target in space, align it in time only for synchronized cameras, and give each exemplar a temporal position but no location in the target.

\paragraph{Reference-guided video editing and object insertion.}
Early diffusion-based video editors follow a text prompt, a reference image or an edited first frame~\cite{tuneavideo,fatezero,tokenflow,gen1,anyv2v}.
Unified editors such as VACE~\cite{vace}, UNIC~\cite{unic}, UniVideo~\cite{univideo} and Kiwi-Edit~\cite{kiwiedit} also take reference images.
Commercial and open-weight systems edit a video with reference images, as Kling~3.0 Omni~\cite{kling3omni} does, or with reference videos, as Wan~3.0~\cite{wan3}, MiniMax-H3~\cite{minimaxh3} and Seedance~2.0~\cite{seedance2} do.
Dedicated insertion methods place the object with boxes and trajectories~\cite{videoanydoor,gencompositor}, reference images with text or an optional mask~\cite{getinvideo,omniinsert}, a 3D pose anchored in one frame~\cite{insertanywhere} or reference views lifted to multiple views~\cite{movi}.
Image-level counterparts insert an exemplar into a masked or described region~\cite{paintbyexample,anydoor,insertanything}, add a described object at a plausible place~\cite{addit}, or render a reference subject with fine details~\cite{fineforge}.
In portrait animation, reference recordings of the same person ground fine details of the mouth, such as the teeth~\cite{rgor}.
Across these methods, the reference shows what the object looks like.
Where it goes is given by a mask, a box, a trajectory, a 3D placement, an edited frame or the text.
In contrast, our reference is another shot of the same scene, and \ours reads the location of the object from it across viewpoints.

\paragraph{Multi-shot and multi-view consistency.}
Multi-shot generation models produce several shots of a scene with shared characters and settings~\cite{lct,shotadapter,captaincinema,moc,holocine,pecontext,multishotmaster,onestory}.
Recent work extends editing to multi-shot videos.
MSEditor~\cite{mseditor} propagates an edit of the first frame across shots, and Thinking on Shots~\cite{thinkingonshots} edits long multi-shot videos with an agent.
Runway Aleph~2.0~\cite{aleph2} edits all shots of a multi-cut video, and ContextMaster~\cite{contextmaster} keeps earlier shots in context.
Group Editing~\cite{groupediting} edits related images with geometric correspondences.
For images, reference-guided inpainting fills a masked region with content from photos of the same scene taken from other viewpoints~\cite{transfill,realfill,leftrefill}, and MVInpainter~\cite{mvinpainter} spreads an edited view to other views.
Camera-controlled re-rendering methods~\cite{recammaster,trajectorycrafter} synthesize a given video from a new camera trajectory.
Multi-view video generators render one scene from several cameras given their poses~\cite{syncammaster,cvd}.
On synchronized recordings such as Ego-Exo4D~\cite{egoexo4d}, cross-view generation translates a video from one camera to another, either with camera poses and rendered geometry~\cite{exo2egov,egox} or in context without them~\cite{egoexogen,worldwander}.
These methods propagate an edit that the user specifies, need a mask, work on images, or generate a whole video from a new camera.
In contrast, \ours inserts an object into an existing shot and reads its location from another shot, without a mask.

{\color{black}
\paragraph{Geometry-aware attention.}
Epipolar constraints guide attention in multi-view feature fusion and reconstruction~\cite{epipolartransformers,gpnr,pixelsplat} and in multi-view and camera-controlled diffusion~\cite{poseguided,epidiff,spad,camco,cvd}.
These methods compute the epipolar lines from known camera poses.
Without poses at inference, epipolar lines can supervise cross-attention during training~\cite{lighttouch}, and matchers fit the geometry to weighted matching points~\cite{goodcorr,deepfundamental,sem,gafm}.
Correspondences also emerge in diffusion features and can guide or supervise generation~\cite{dift,difftrack,corrfill,track4gen,cameo,mvtrack4gen}.
In contrast, \ours estimates the epipolar geometry between two shots from its own matching points at inference, without camera poses, and uses it to bias its attention toward the epipolar lines and the matching exemplars.
}

\begin{figure*}[t]
\centering
\includegraphics[width=\textwidth]{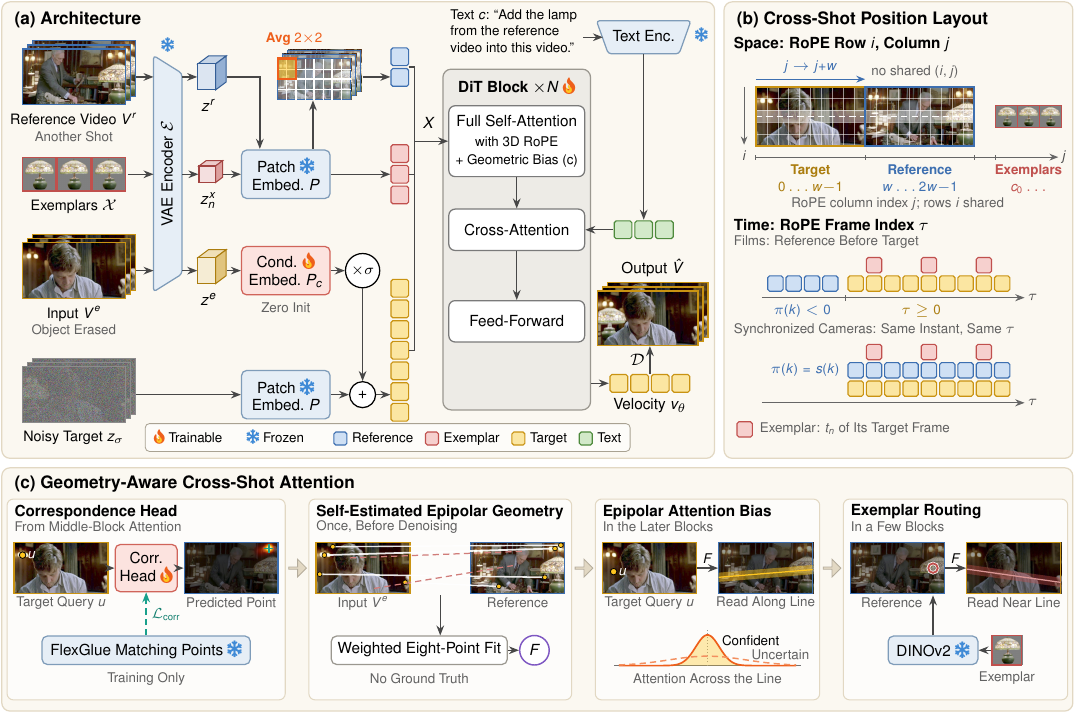}
\caption{Overview of \ours. (a) The reference, exemplar and noisy target tokens form one sequence, and the input video is added to the target tokens. (b) Positions of the reference and exemplar tokens. (c) Epipolar geometry estimated from the model's own matching points guides the attention to the reference and the exemplars.}
\label{fig:method}
\end{figure*}

%% file: sec/3_method.tex
\section{Method}
\label{sec:method}

\subsection{Problem Formulation}
\label{sec:problem}

We study cross-shot video object insertion as a conditional video generation task.
The input video $V^{e}$ is a target shot of a scene from which an object is absent, and $V$ denotes the same shot with the object, the ground truth during training.
The reference video $V^{r}$ is a short clip from another shot of the same scene in which the object is visible.
The object exemplars $\mathcal{X}=\{(x_n,t_n)\}_{n=1}^{N}$ are images of the object alone on a plain gray background, where $x_n$ shows how the object should look from the viewpoint of the target shot at frame $t_n$; such views are easy to obtain by simply taking a photo of the object from that angle, and several objects can be inserted at once by giving the exemplars of each.
The text prompt $c$ names the object and states how the two videos were recorded.
Given $(V^{e}, V^{r}, \mathcal{X}, c)$ and no mask, box or camera parameters, the goal is an output $\hat{V}$ with four properties: \textbf{geometric consistency}, the object stands at its place in the shared 3D scene and is seen with the scale, perspective and occlusion of the target camera (\cref{sec:geometry});
\textbf{appearance fidelity}, it keeps the details of the exemplars and the reference;
\textbf{temporal consistency}, it stays stable over time;
and \textbf{preservation}, the rest of the shot stays unchanged.

\subsection{Overall Architecture}
\label{sec:arch}

\ours builds on the DiT video model Wan2.2-TI2V-5B~\cite{wan}, adapted with LoRA~\cite{lora} (\cref{fig:method}(a)).
Its VAE encoder $\mathcal{E}$ maps the videos and the exemplars to latents $z_0=\mathcal{E}(V)$, $z^{e}=\mathcal{E}(V^{e})$, $z^{r}=\mathcal{E}(V^{r})$ and $z^{x}_n=\mathcal{E}(x_n)$.
The noisy target latent at noise level $\sigma\in[0,1]$ is
\begin{equation}
z_\sigma=(1-\sigma)\,z_0+\sigma\,\epsilon,\qquad \epsilon\sim\mathcal{N}(0,I).
\label{eq:noisy}
\end{equation}
The input video is summed into the target tokens~\cite{kiwiedit},
\begin{equation}
X^{t}=P(z_\sigma)+\sigma\,P_c(z^{e}),
\label{eq:cond}
\end{equation}
where $P$ is the pretrained patch embedding and $P_c$ is a new embedding of the same shape, initialized to zero.
We arrange the reference tokens by frame, row and column and average every $2\times2$ block within a frame,
{\color{black}
\begin{equation}
X^{r}_{k,I,J}=\tfrac{1}{4}\textstyle\sum_{a,b\in\{0,1\}}P(z^{r})_{k,\,2I+a,\,2J+b},
\label{eq:pool}
\end{equation}}%
where $k$ indexes the reference frames, $(I,J)$ indexes a block and $a,b\in\{0,1\}$ index the four tokens in it, which keeps a quarter of the tokens.
We concatenate all tokens,
\begin{equation}
X=\big[\,X^{r}\ \Vert\ X^{x}\ \Vert\ X^{t}\,\big],
\label{eq:concat}
\end{equation}
and the DiT processes them with full self-attention and with cross-attention to the text $c$.

\subsection{Cross-Shot Position Layout}
\label{sec:layout}

The DiT encodes the frame $\tau$, row $i$ and column $j$ of each token with 3D RoPE~\cite{rope}.
Tokens of another camera at the same $(i,j)$ do not show the same place, so we shift the column index of each pooled reference token, taken at the center of its block, by the width $w$ of the target grid, as if the two shots were shown side by side (\cref{fig:method}(b)).
The exemplars form a further panel.
In time, the reference frames precede the target frames for film shots and share their indices for synchronized cameras, and each exemplar takes the index of its target frame $t_n$~\cite{paistudio}.

{\color{black}
\subsection{Geometry-Aware Cross-Shot Attention}
\label{sec:geometry}

The layout does not tell a target token where to look in the reference, and by appearance alone it can pick a similar region elsewhere in the scene.
We therefore make the attention follow the geometry of the two cameras.
For two cameras viewing the same scene, a point seen at $x$ in one image appears on the epipolar line $F\tilde{x}$ in the other, where $F$ is the $3\times3$ fundamental matrix~\cite{hartley8pt} and $\tilde{x}$ is $x$ in homogeneous coordinates.
\ours estimates $F$ itself (\cref{fig:method}(c)): a correspondence head finds matching points between the two videos, a weighted fit turns them into $F$, and two attention biases make the target follow the epipolar lines.

\begin{figure}[t]
\centering
\includegraphics[width=\linewidth]{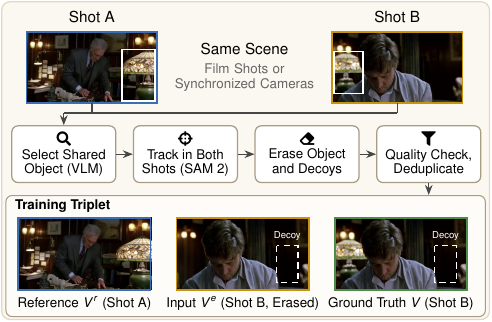}
\caption{Erase-and-restore data engine.}
\label{fig:data}
\end{figure}

\paragraph{Correspondence head.}
For a target token $u$ and a pooled reference token $v$, the head combines the query-key scores of all heads $h$ in a group of intermediate blocks $\mathcal{B}$,
\begin{equation}
A(u,v)=\frac{1}{T}\sum_{b\in\mathcal{B}}\sum_{h}a_{b,h}\,\frac{q_{b,h}(u)^{\top}k_{b,h}(v)}{\sqrt{d_h}},
\label{eq:corr}
\end{equation}
where the head weights $a_{b,h}\geq0$ are learned and sum to one, $d_h$ is the head dimension and $T$ is a temperature.
Within one reference frame, $p_u=\mathrm{softmax}_v A(u,v)$ gives the probability that $v$ shows the same point of the scene as $u$.
During training, we supervise $p_u$ with matching points between the ground-truth target frames and the reference frames, found offline by FlexGlue~\cite{flexglue} and filtered by MAGSAC++~\cite{magsacpp},
\begin{equation}
\mathcal{L}_{\mathrm{corr}}=\frac{\sum_i \omega_i\,\mathrm{CE}\big(p_{u_i},\,G_i\big)}{\sum_i \omega_i},
\label{eq:lcorr}
\end{equation}
where pair $i$ links the target token $u_i$ to a reference location, $G_i$ is a small Gaussian around it and $\omega_i$ is its confidence.
$\mathcal{L}_{\mathrm{corr}}$ updates only $a$ and the queries and keys of $\mathcal{B}$.

\paragraph{Self-estimated epipolar geometry.}
Before denoising, \ours runs its first blocks once on the input video and the reference.
For probe tokens $u$, the mean position under $p_u$ gives a matching point and the spread of $p_u$ gives its certainty.
A weighted eight-point algorithm~\cite{hartley8pt} fits $F$ to these points, and three reweighting rounds lower the weight of points far from their epipolar lines.
The confidence $c\in[0,1]$ is high when the points are certain and agree with $F$.

\paragraph{Epipolar attention bias.}
In the later blocks, the attention logit between $u$ and $v$ receives
\begin{equation}
\beta(u,v)=-\alpha_{b,h}\,c\,d(u,v)^{2}+\log Z_u,
\label{eq:epibias}
\end{equation}
where $d(u,v)$ is the distance from $u$ to the epipolar line of $v$, $\alpha_{b,h}\geq0$ is learned per block and head, and $Z_u$ keeps the total attention of $u$ to the reference unchanged.
Each target token thus reads the reference along its epipolar line, and the bias vanishes when $c=0$.

\paragraph{Exemplar routing.}
The same geometry also tells where the object must appear.
DINOv2~\cite{dinov2} locates the object of exemplar $n$ in the reference at $o_n$, and a fundamental matrix $F_\tau$, fitted with RANSAC~\cite{ransac} to the model's own matching points of target frame $\tau$, maps it to the line $\ell_{n,\tau}=F_\tau\tilde{o}_n$.
In a few blocks, the logit between $u$ and exemplar $n$ receives
\begin{equation}
\gamma(u,n)=-\mu_{b,h}\,c_{n,\tau}\,\bar{d}_{n}(u)^{2}+\log Z_{n,\tau},
\label{eq:route}
\end{equation}
where $\bar{d}_{n}(u)$ is the distance from $u$ to a band around $\ell_{n,\tau}$, $c_{n,\tau}$ is a confidence from the fit and DINOv2, $\mu_{b,h}\geq0$ is learned, and $Z_{n,\tau}$ keeps the total exemplar attention unchanged, so tokens near the line read the exemplars more.%
}

\subsection{Training Objective}
\label{sec:objective}

We train the LoRA weights, $P_c$, $a$, $\alpha$ and $\mu$ so that the velocity $v_\theta$ predicted at the target tokens matches $v=\epsilon-z_0$, with a larger weight $\lambda>1$ on the cells of the object region $m$, since the object is usually small,
\begin{equation}
\mathcal{L}_{\mathrm{FM}}=\frac{1}{|\Omega|}\sum_{\Omega}\big(1+(\lambda-1)\,m\big)\odot\lVert v_\theta-v\rVert^{2},
\label{eq:loss}
\end{equation}
where $\Omega$ is the set of latent cells of the target.
At low noise levels we add $\mathcal{L}_{\mathrm{perc}}$, the LPIPS distance~\cite{lpips} between the decoded prediction and the ground truth in the object region together with the difference of their frame-to-frame changes, and minimize
{\color{black}
\begin{equation}
\mathcal{L}=\mathcal{L}_{\mathrm{FM}}+\eta_{c}\,\mathcal{L}_{\mathrm{corr}}+\eta_{p}\,\mathcal{L}_{\mathrm{perc}}.
\label{eq:total}
\end{equation}}%
Here $\eta_{c}$ and $\eta_{p}$ weight the two additional losses, and the mask $m$ is used only in the losses.

\begin{table*}[t]
\centering
\caption{Quantitative comparison on the benchmark. Object fidelity is measured in a crop around the object, and the whole-video metrics are measured over the full frame, both against the ground truth (\cref{sec:setup}); MSE is given in units of $10^{-3}$. The best result is in bold, and the second best is underlined.}
\label{tab:main}
\scriptsize
\setlength{\tabcolsep}{0pt}
\renewcommand{\arraystretch}{1.0}
\begin{tabularx}{\textwidth}{@{}l*{8}{>{\centering\arraybackslash}X}@{}}
\toprule
& \multicolumn{2}{c}{Placement} & \multicolumn{3}{c}{Object fidelity} & \multicolumn{3}{c}{Whole video} \\
\cmidrule(lr){2-3}\cmidrule(lr){4-6}\cmidrule(lr){7-9}
Method & IoU$\uparrow$ & Cen.$\downarrow$ & CLIP-I$\uparrow$ & DINO-I$\uparrow$ & LPIPS$\downarrow$ & MSE$\downarrow$ & SSIM$\uparrow$ & FVD$\downarrow$ \\
\midrule
VACE-14B & 0.0184 & 0.200 & 0.871 & 0.729 & 0.354 & 9.86 & 0.728 & 330 \\
UniVideo & 0.0685 & 0.107 & 0.872 & 0.702 & 0.432 & 28.3 & 0.524 & 415 \\
Kiwi-Edit & 0.426 & 0.0472 & 0.915 & 0.801 & 0.270 & 4.56 & \underline{0.876} & \underline{157} \\
MiniMax-H3 & 0.125 & 0.155 & 0.888 & 0.733 & 0.383 & 31.6 & 0.568 & 398 \\
Kling 3.0 Omni & 0.234 & 0.113 & 0.876 & 0.713 & 0.393 & 26.9 & 0.689 & 384 \\
Wan 3.0 & \underline{0.572} & \underline{0.0287} & \underline{0.931} & \underline{0.891} & \underline{0.192} & \underline{4.26} & 0.779 & 161 \\
\rowcolor{oursrow}
\textbf{\ours (ours)} & \textbf{0.674} & \textbf{0.0237} & \textbf{0.952} & \textbf{0.926} & \textbf{0.132} & \textbf{1.51} & \textbf{0.905} & \textbf{74.9} \\
\bottomrule
\end{tabularx}
\end{table*}

\begin{figure*}[t]
\centering
\begin{animateinline}[poster=first,width=\textwidth]{100}
\includegraphics[width=\textwidth]{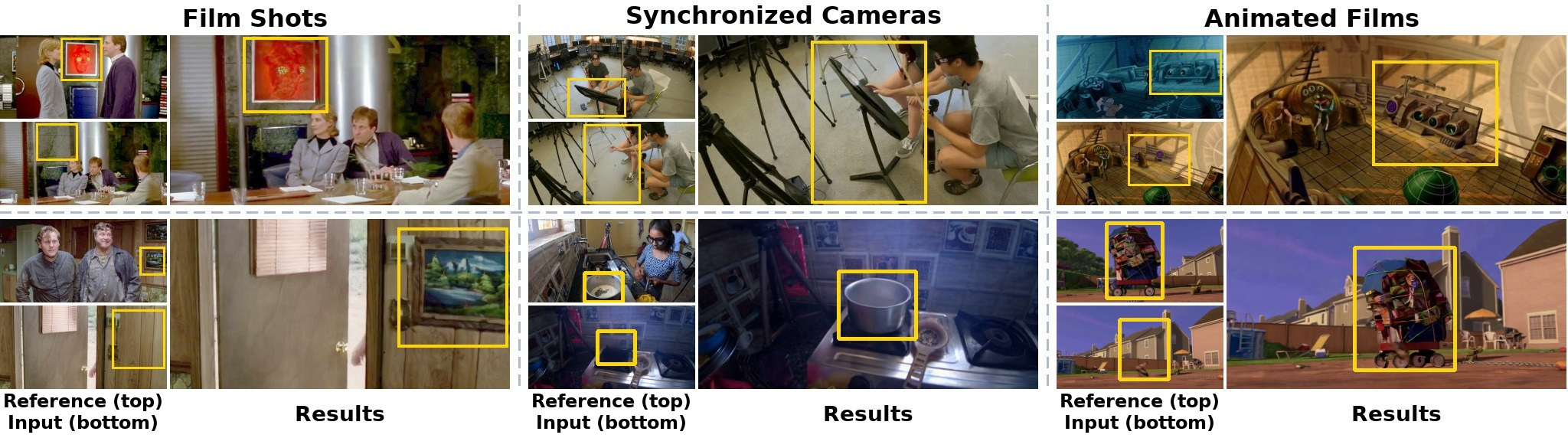}
\newframe[2]
\multiframe{8}{i=0+1}{\includegraphics[width=\textwidth]{figures/showcase/f_\i.jpg}}
\end{animateinline}
\caption{Results of \ours on film shots, synchronized cameras and animated films. Yellow boxes mark the object in the reference video, the input video and the result. More results are in \mbox{\cref{supp:results}}. Readers can click and play the video clips in this figure using \textcolor{red}{\textbf{Adobe Acrobat}}.}
\label{fig:showcase}
\end{figure*}

\subsection{Erase-and-Restore Data Engine}
\label{sec:data}

We obtain training samples by erasing an object from real footage (\cref{fig:data}).
In feature films, a vision-language model~\cite{gemini} selects an object that two shots of a scene both show, SAM~2~\cite{sam2} tracks it, and an image editor~\cite{qwenimage} with a video object remover~\cite{minimaxremover} erases it from one shot, which becomes the input video; its original is the ground truth and the other shot is the reference.
From Ego-Exo4D~\cite{egoexo4d}, we pair two synchronized views and erase the object in the same way.
Random decoy regions are erased in both input and ground truth.
The exemplars are cut from the ground-truth shot and randomly perturbed in color, rotation and aspect ratio during training, and the FlexGlue matching points and DINOv2 features are computed once, offline.
The training set has 7.99k film and 9.85k Ego-Exo4D clips, and the benchmark has 100 clips (\mbox{\cref{sec:setup}}). \mbox{\cref{tab:supp_stats}} gives the statistics of both sets, and \mbox{\cref{supp:data}} details the data engine.

%% file: sec/4_experiments.tex
\section{Experiments}
\label{sec:experiments}

\subsection{Experimental Setup}
\label{sec:setup}

\paragraph{Implementation details.}
We train LoRA~\cite{lora} adapters of rank 80 on Wan2.2-TI2V-5B~\cite{wan}, together with $P_c$, the head weights $a$ and the bias strengths $\alpha$ and $\mu$.
We set $\eta_c=10^{-4}$ and choose $\eta_p$ at every step so that the gradient of $\mathcal{L}_{\mathrm{perc}}$ on the object region has half the norm of that of $\mathcal{L}_{\mathrm{FM}}$~\cite{vqgan}.
The reference has up to 33 frames and the target has 49, 65 or 81 frames, each clip comes with six exemplars, and inference uses an exponential moving average of the trained weights.

\paragraph{Baselines.}
We compare with six reference-guided video editors: three open-source editors (VACE-14B~\cite{vace}, UniVideo~\cite{univideo} and Kiwi-Edit~\cite{kiwiedit}) and three commercial models (MiniMax-H3~\cite{minimaxh3}, Kling~3.0 Omni~\cite{kling3omni} and Wan~3.0~\cite{wan3}).
MiniMax-H3 and Wan~3.0 also accept a reference video and are the closest to our setting.
Every method runs in its best setting (more details in \cref{supp:metrics}).

\paragraph{Benchmark.}
Our benchmark contains 100 clips randomly sampled from the output of the data engine: 60 pairs of shots from feature films and 40 camera pairs from Ego-Exo4D~\cite{egoexo4d}.
The film clips come from 57 feature films released between 1960 and 2006, including live-action films in color and in black and white, and 2D and 3D animated films.
The Ego-Exo4D clips cover cooking, bike repair, COVID tests, music and ball sports, with hand-held and moving objects.

\paragraph{Evaluation metrics.}
(1) Placement: \textbf{IoU} between the region that a method changes in the input video and the region where the ground truth differs from it, and \textbf{Cen.}, the distance between their centers over the frame diagonal.
(2) Object fidelity in a crop around the object: \textbf{CLIP-I} and \textbf{DINO-I}, the cosine similarity of CLIP~\cite{clip} and DINO~\cite{dino} features~\cite{dreambooth}, and \textbf{LPIPS}~\cite{lpips}.
(3) Whole video: \textbf{MSE}, \textbf{SSIM}~\cite{ssim} and \textbf{FVD}~\cite{fvd}.
All metrics compare with the ground truth. IoU and Cen.\ take the median over the frames of a clip, and the other frame-level metrics take the mean. \cref{supp:metrics} gives the exact definitions.

\begin{figure*}[t]
\centering
\begin{animateinline}[poster=first,width=0.97\textwidth]{100}
\includegraphics[width=0.97\textwidth]{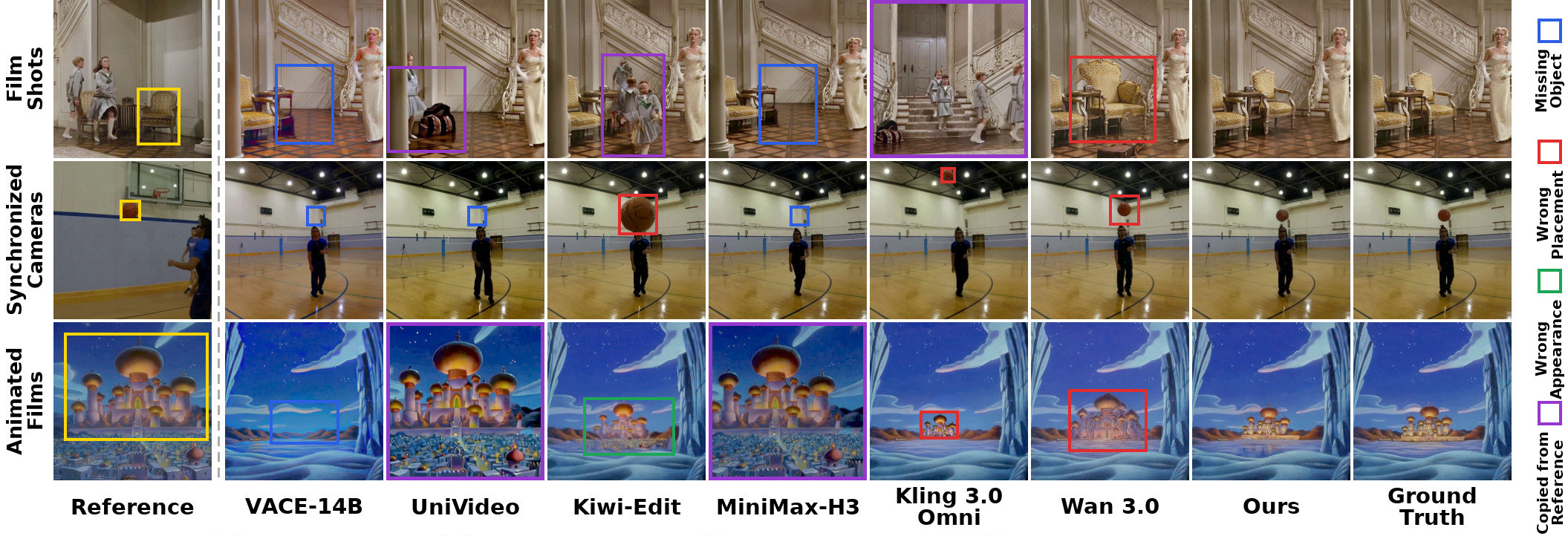}
\newframe[2]
\multiframe{8}{i=0+1}{\includegraphics[width=0.97\textwidth]{figures/comparison/f_\i.jpg}}
\end{animateinline}
\caption{Comparison with six video editors on film shots, synchronized cameras and animated films; the yellow box marks the object in the reference, and colored boxes mark the failures. Readers can click and play the video clips in this figure using \textcolor{red}{\textbf{Adobe Acrobat}}.}
\label{fig:qualitative}
\end{figure*}

\subsection{Comparison and Evaluation}
\label{sec:comparison}
\label{sec:results}

\paragraph{Quantitative evaluation.}
\begin{table}[t]
\centering
\caption{Cost of editing one benchmark clip, averaged over the clips. The cost of \ours is based on inference on one RTX 5090, and the cost of the commercial editors is based on their public API prices.}
\label{tab:cost}
\scriptsize
\setlength{\tabcolsep}{4pt}
\renewcommand{\arraystretch}{1.0}
\begin{tabularx}{\linewidth}{@{}l>{\raggedright\arraybackslash}X r@{}}
\toprule
Method & Price basis & Cost (\$) \\
\midrule
MiniMax-H3 & \$0.06 per second of output & 0.300 \\
Kling 3.0 Omni & \$0.168 per second of output & 0.536 \\
Wan 3.0 & \$0.10 per second of input and output & 0.732 \\
\rowcolor{oursrow}
\textbf{\ours (ours)} & \$0.556 per GPU-hour, 167\,s per clip & \textbf{0.0258} \\
\bottomrule
\end{tabularx}
\end{table}
\ours achieves the best result on every metric of \mbox{\cref{tab:main}}.
\textbf{Placement.}
Guided by the epipolar geometry between the input video and the reference video, \ours places objects closest to their true location, with an IoU of 0.674 against 0.572 for Wan~3.0, the strongest baseline, and the lowest center error.
\textbf{Object fidelity.}
The inserted objects also match the ground truth most closely in shape, color and detail (DINO-I 0.926 against 0.891 for Wan~3.0), since the exemplars show them from the target viewpoint and exemplar routing directs their appearance to the target tokens along the epipolar line of the object.
\textbf{Whole video.}
Over the full frame, \ours has the lowest MSE and the highest SSIM (MSE 1.51 against 4.26 for the next-best editor), and its videos have the lowest FVD (74.9 against 157).
Because the input video is summed into the target tokens, \ours keeps the rest of the shot intact, whereas several editors regenerate large parts of it.

\paragraph{Visual comparison.}
\cref{fig:showcase} shows results of \ours on film shots, synchronized cameras and animated films.
With the geometry-aware attention, \ours places the objects where the reference shows them and at the scale of the target camera.
Their shape and texture follow the exemplars, while their lighting and style follow the target shot, as if the objects had been filmed with the scene.
They stay consistent over time as the camera and the people move, and the rest of the shot remains unchanged.
In \mbox{\cref{fig:qualitative}}, existing editors commonly omit the objects, copy the view or the content of the reference shot into the target, or draw them at a wrong place, scale or shape, as highlighted by the boxes.
In contrast, \ours reads the location from the reference video and places the objects where they stand in the target shot, with the viewpoint, scale and occlusion of the target camera, and leaves the rest of the shot unchanged.
Notably, it remains accurate in challenging cases where baselines typically fail, such as a distant object seen from a very different angle or a ball in flight.
More results and comparisons are in \cref{supp:results}.

\begin{table*}[t]
\centering
\caption{Ablation on the benchmark, with the metrics of \cref{tab:main}. The best result is in bold.}
\label{tab:ablation}
\scriptsize
\setlength{\tabcolsep}{0pt}
\renewcommand{\arraystretch}{1.0}
\begin{tabularx}{\textwidth}{@{}l@{\hspace{2pt}}l@{\hspace{2pt}}l*{8}{>{\centering\arraybackslash}X}@{}}
\toprule
\multicolumn{3}{@{}l}{} & \multicolumn{2}{c}{Placement} & \multicolumn{3}{c}{Object fidelity} & \multicolumn{3}{c}{Whole video} \\
\cmidrule(lr){4-5}\cmidrule(lr){6-8}\cmidrule(lr){9-11}
\multicolumn{3}{@{}l}{Variant} & IoU$\uparrow$ & Cen.$\downarrow$ & CLIP-I$\uparrow$ & DINO-I$\uparrow$ & LPIPS$\downarrow$ & MSE$\downarrow$ & SSIM$\uparrow$ & FVD$\downarrow$ \\
\midrule
\multicolumn{3}{@{}l}{w/o geometry-aware attention} & 0.570 & 0.0276 & 0.941 & 0.905 & 0.170 & 2.37 & 0.877 & 116 \\
\multirow{4}{*}{Number of exemplars} & \ldelim\{{4}{3mm} & 5 & 0.664 & 0.0272 & 0.950 & 0.921 & 0.133 & 1.55 & 0.905 & 80.0 \\
& & 3 & 0.660 & 0.0259 & 0.950 & 0.922 & 0.135 & 1.54 & 0.905 & 77.2 \\
& & 1 & 0.621 & 0.0261 & 0.944 & 0.908 & 0.152 & 1.78 & 0.903 & 83.6 \\
& & 0 & 0.552 & 0.0242 & 0.928 & 0.844 & 0.204 & 2.44 & 0.899 & 119 \\
\multicolumn{3}{@{}l}{3D views instead of exemplars} & 0.547 & 0.0270 & 0.926 & 0.847 & 0.203 & 2.28 & 0.900 & 105 \\
\rowcolor{oursrow}
\multicolumn{3}{@{}l}{\textbf{Full Model}} & \textbf{0.674} & \textbf{0.0237} & \textbf{0.952} & \textbf{0.926} & \textbf{0.132} & \textbf{1.51} & \textbf{0.905} & \textbf{74.9} \\
\bottomrule
\end{tabularx}
\end{table*}

\begin{figure*}[t]
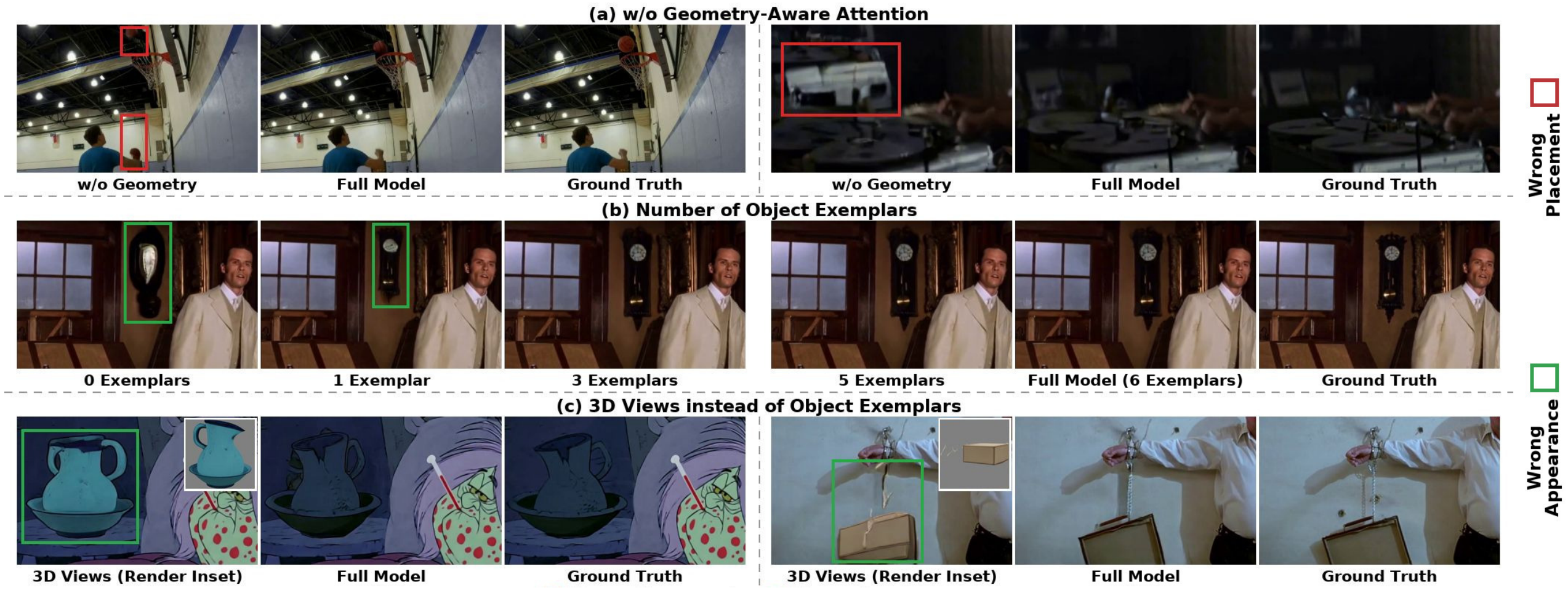

\centering
\animategraphics[width=0.96\textwidth,poster=3]{2}{figures/ablation/f_}{0}{7}
\caption{Ablation study: (a) without geometry-aware attention, (b) with different numbers of object exemplars and (c) with 3D views instead of exemplars (inset: the 3D render it receives). Readers can click and play the video clips in this figure using \textcolor{red}{\textbf{Adobe Acrobat}}.}
\label{fig:ablation}
\end{figure*}

\paragraph{Efficiency.}
Pooling the reference makes \ours faster than the same backbone fed the full reference, although it adds the geometry-aware attention.
\ours is trained and evaluated with 33-frame references, with which it generates a clip 1.30 times faster.
To see how the cost scales, we also time generation with references of 65, 97 and 129 frames: the gap grows with the reference length, to 2.25 times at 129 frames (\mbox{\cref{supp:cost}}).
One clip costs \$0.0258 on one RTX 5090, over ten times less than any commercial editor (\cref{tab:cost}).

\subsection{Ablation Study}
\label{sec:ablation}
\begin{figure}[t]
\centering
\includegraphics[width=\linewidth]{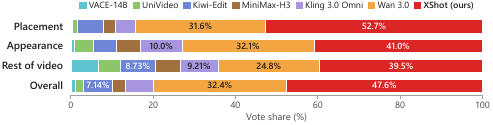}
\caption{User study: share of votes received by each method for each criterion. \ours receives the most votes on all four criteria.}
\label{fig:userstudy}
\end{figure}

We conduct ablation experiments on the benchmark (\cref{tab:ablation}), where the full \ours is best on every metric.

\textbf{Without geometry-aware attention.}
Removing the geometry-aware attention lowers IoU from 0.674 to 0.570 and raises LPIPS from 0.132 to 0.170.
Without the epipolar geometry between the two shots, the model has to infer from appearance alone where a point of the reference shot lands in the target, so it places objects less accurately and renders them less faithfully, for example at a wrong place or scale (\cref{fig:ablation}(a)), and it also changes more of the background.
These results demonstrate that the geometry-aware attention is crucial for both accurate placement and faithful rendering of the inserted objects.

\textbf{Fewer object exemplars.} Each clip comes with six exemplars, shared evenly among the inserted objects. Keeping five or three of them gives results close to the full model, although with several objects in a clip each object then keeps even fewer exemplars, which can limit how faithfully it is rendered. Keeping one lowers IoU from 0.674 to 0.621 and raises LPIPS from 0.132 to 0.152, most when the camera moves strongly, since the target shot then shows the object from viewpoints that a single exemplar does not cover. Without any exemplar, the model has to infer by itself how the object looks from the viewpoints of the target shot, which is hard: it still finds where the object stands but renders it much less faithfully (\mbox{\cref{fig:ablation}(b)}).

\textbf{3D views instead of exemplars.}
We also explored providing other views of the object as orbit renders of its 3D reconstruction~\cite{sam3d} built from the reference shot.
This variant places and renders the object worse, about as well as without any exemplars, since 3D reconstructions from real footage are coarse and lose many object details (\mbox{\cref{fig:ablation}(c)}).

\subsection{User Study}
\label{sec:userstudy}

We conduct a user study to evaluate the results from a human perspective on 30 clips randomly sampled from the benchmark.
For each clip, participants see the reference shot, the input video and the ground truth at the top and the anonymized outputs of all seven methods below, in a random order.
They choose the best output for each of four criteria: object placement, object appearance, preservation of the rest of the video and overall realism.
We report the votes of 21 participants (details in \cref{supp:userstudy}).
As shown in \cref{fig:userstudy}, \ours is consistently preferred across all four criteria, receiving about half of all votes for placement and overall realism, which confirms that its insertions are also the most convincing to human viewers.

%% file: sec/5_conclusion.tex
\section{Conclusion}
\label{sec:conclusion}

We introduced the new task of cross-shot video object insertion and presented \ours, which reads where objects stand from a reference video of another shot and takes their appearance from exemplars.
Its geometry-aware cross-shot attention estimates the epipolar geometry between the two shots from its own matching points and lets objects land at the right place and with the right size.
Our erase-and-restore data engine turns films and multi-camera recordings into training data and a benchmark, on which \ours places objects more accurately and renders them more faithfully than open-source and commercial video editors.

%% file: sec/6_llm_statement.tex
\section*{LLM Usage Statement}
We used Claude to help write and debug code for training, evaluation and the data engine, to help prepare figures, and to polish the writing.
Our data engine (\cref{sec:data}) uses Gemini to select and localize objects, and Qwen-Image and MiniMax-Remover to erase them.
One ablation uses SAM~3D to reconstruct objects.
The authors reviewed all AI-assisted code, figures and text and take full responsibility for this paper.

%% file: sec/X_suppl.tex
\clearpage
\makeatletter\renewcommand*{\Hy@backout}[1]{}\makeatother
\newpage\twocolumn[\centering\Large\textbf{\thetitle}\\\vspace{0.5em}Appendix\\\vspace{1.0em}]
\appendix
\renewcommand{\thesection}{S\arabic{section}}
\setcounter{table}{0}
\renewcommand{\thetable}{S\arabic{table}}
\setcounter{figure}{0}
\renewcommand{\thefigure}{S\arabic{figure}}
\setcounter{equation}{0}
\renewcommand{\theequation}{S\arabic{equation}}
\renewcommand{\theHequation}{S\arabic{equation}}
\setcounter{dbltopnumber}{1}
\renewcommand{\dbltopfraction}{0.9}
\renewcommand{\dblfloatpagefraction}{0.6}

The appendix is organized as follows:
\begin{enumerate}[nosep]
\item Sec.~\ref{supp:userstudy}: interface, protocol and questions of the user study, and its correlation with the metrics.
\item Sec.~\ref{supp:metrics}: definitions of the evaluation metrics.
\item Sec.~\ref{supp:results}: additional results and comparisons.
\item Sec.~\ref{supp:data}: training data and the instruction of the vision-language model.
\item Sec.~\ref{supp:geometry}: details of the geometry-aware cross-shot attention.
\item Sec.~\ref{supp:cost}: speed for longer references and cost against commercial editors.
\item Sec.~\ref{supp:failures}: failure cases, limitations and future work.
\item Sec.~\ref{supp:ethics}: ethics statement.
\end{enumerate}

\section{User Study Details}
\label{supp:userstudy}

\begin{figure*}[t]
\centering
\includegraphics[width=\textwidth]{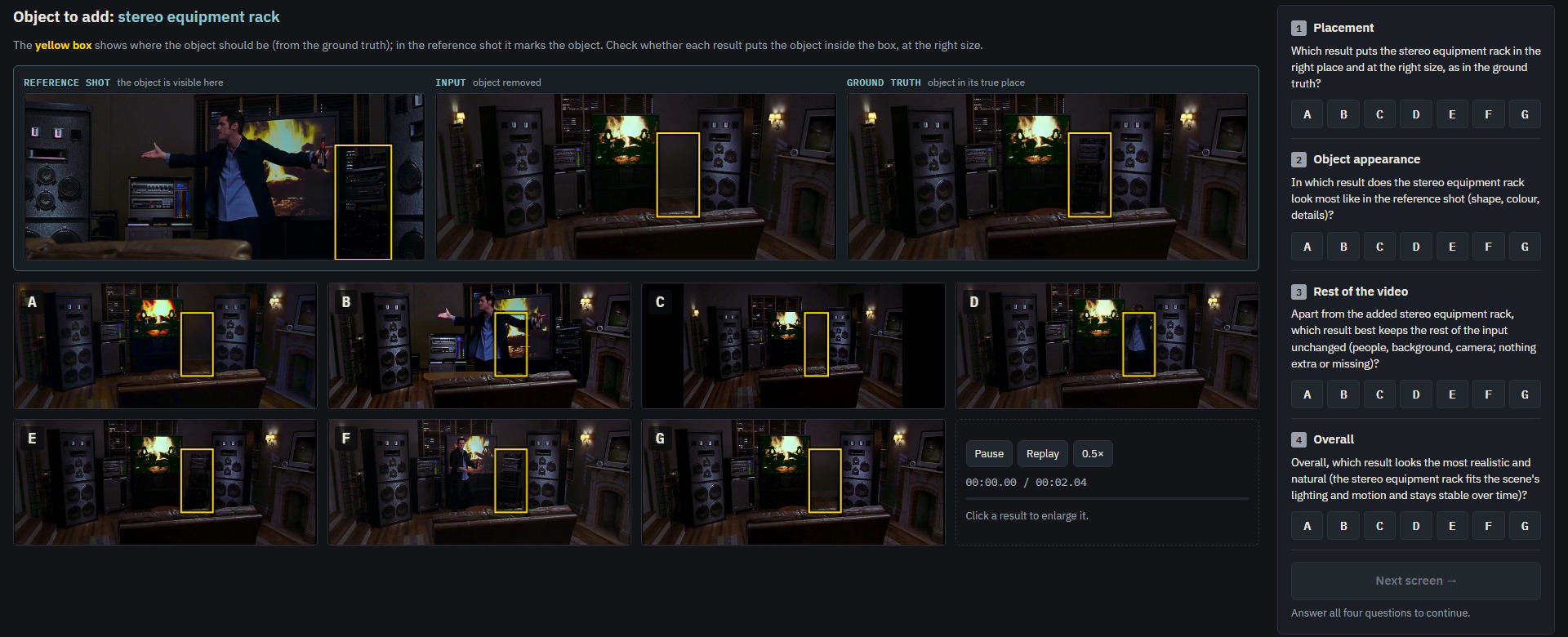}
\caption{User study interface. Top: the reference shot, the input video and the ground truth, with yellow boxes marking the object in the reference and where it should appear in every other video. Bottom: the seven anonymized outputs, playing in sync. Right: the four questions asked for each clip.}
\label{fig:supp_userstudy}
\end{figure*}

\paragraph{Protocol.}
\cref{fig:supp_userstudy} shows the interface.
We randomly sample 30 clips from the benchmark.
Each participant judges all 30 clips in a random order.
The reference shot, the input video and the ground truth are shown at the top; a yellow box computed from the ground truth marks the object in the reference shot and the place where it should appear in every other video, including all seven outputs.
The outputs are labeled A to G in an order randomized for every participant and clip.
All videos play in sync and can be paused, replayed, slowed down and enlarged.
For each clip, participants answer four single-choice questions, where $\langle$object$\rangle$ is the object named in the instruction:
\begin{itemize}[nosep]
\item Placement: ``Which result puts the $\langle$object$\rangle$ in the right place and at the right size, as in the ground truth?''
\item Appearance: ``In which result does the $\langle$object$\rangle$ look most like in the reference shot (shape, colour, details)?''
\item Rest of the video: ``Apart from the added $\langle$object$\rangle$, which result best keeps the rest of the input unchanged (people, background, camera; nothing extra or missing)?''
\item Overall: ``Overall, which result looks the most realistic and natural (the $\langle$object$\rangle$ fits the scene's lighting and motion and stays stable over time)?''
\end{itemize}

\paragraph{Participants.}
We inspect every submission and discard those that repeatedly chose an output that clearly fails the criterion asked, such as one in which the object is missing or the shot is replaced by the reference shot.

\paragraph{Correlation with the automatic metrics.}
For each clip, we compare the share of votes of each method with its automatic score, turned into a distribution over the seven methods by a softmax of the score standardized within the clip, with lower-is-better metrics negated~\cite{paistudio}.
\cref{tab:supp_corr} reports Spearman's $\rho$ over all clip--method pairs of the study.
Every metric correlates positively with every criterion ($p<0.001$), and the object metrics DINO-I and LPIPS and the placement metric IoU agree most with the votes.
FVD has one value per method rather than per clip and is therefore not included.

\begin{table}[!h]
\centering
\caption{Spearman's $\rho$ between the share of votes and each automatic metric over clip--method pairs; the best metric for each criterion is in bold.}
\label{tab:supp_corr}
\footnotesize
\setlength{\tabcolsep}{3pt}
\renewcommand{\arraystretch}{1.08}
\begin{tabularx}{\linewidth}{@{}l *{4}{>{\centering\arraybackslash}X}@{}}
\toprule
Metric & Placement & Appearance & Rest of video & Overall \\
\midrule
IoU$\uparrow$ & 0.752 & \textbf{0.642} & 0.543 & 0.733 \\
Cen.$\downarrow$ & 0.496 & 0.483 & 0.358 & 0.524 \\
CLIP-I$\uparrow$ & 0.706 & 0.522 & 0.652 & 0.692 \\
DINO-I$\uparrow$ & \textbf{0.766} & 0.588 & 0.711 & \textbf{0.754} \\
LPIPS$\downarrow$ & 0.765 & 0.592 & \textbf{0.735} & 0.753 \\
MSE$\downarrow$ & 0.606 & 0.400 & 0.604 & 0.587 \\
SSIM$\uparrow$ & 0.486 & 0.278 & 0.493 & 0.483 \\
\bottomrule
\end{tabularx}
\end{table}

\section{Evaluation Metrics}
\label{supp:metrics}
For an output frame $O$, let $I$ and $G$ be the input and ground-truth frames at the same time, let $C^{o}$ and $C^{g}$ be the pixels where $O$ and $G$ differ from $I$, let $D$ be the decoy regions, and let $B(\cdot)$ be a square crop around the ground-truth object resized to $224\times224$:
\begin{itemize}[nosep]
\item \textbf{IoU} $=\lvert (C^{o}\cap C^{g})\setminus D\rvert\,/\,\lvert (C^{o}\cup C^{g})\setminus D\rvert$.
\item \textbf{Cen.} $=\lVert c^{o}-c^{g}\rVert_2/d$, where $c^{g}$ is the center of the ground-truth object, $c^{o}$ is the center of the changed region nearest to it and $d$ is the frame diagonal.
\item \textbf{CLIP-I} $=\cos\big(\psi(B(O)),\psi(B(G))\big)$ and \textbf{DINO-I} $=\cos\big(\phi(B(O)),\phi(B(G))\big)$, where $\psi$ is the CLIP ViT-B/32 image embedding~\cite{clip} and $\phi$ is the DINO ViT-S/16 class token~\cite{dino}.
\item \textbf{LPIPS} $=\ell\big(B(O),B(G)\big)$, where $\ell$ is the LPIPS distance~\cite{lpips}.
\item \textbf{MSE} $=\lVert O-G\rVert_2^2/\lvert O\rvert$ and \textbf{SSIM}~\cite{ssim} is computed between $O$ and $G$ over the full frame, with pixel values in $[0,1]$.
\item \textbf{FVD}~\cite{fvd} is the Fr\'echet distance between the I3D features of the outputs and of the ground-truth videos, computed on 16 frames per clip.
\end{itemize}
IoU and Cen.\ take the median over the frames of a clip, and the other frame metrics take the mean; these are averaged over the clips.
Outputs with another frame rate are matched to the input frames by time.

\cref{tab:supp_baselines_ex} lists the inputs of every method. Unless stated otherwise, each method receives six exemplars. With a video, Kling 3.0 Omni accepts at most four images and elements, so it receives three reference frames and one element of four exemplars. Kiwi-Edit only accepts a single reference image.

\begin{table}[!h]
\centering
\caption{Inputs of each method in the test where the other methods also receive the exemplars.}
\label{tab:supp_baselines_ex}
\footnotesize
\setlength{\tabcolsep}{3pt}
\renewcommand{\arraystretch}{1.12}
\begin{tabularx}{\linewidth}{@{}l >{\centering\arraybackslash}X c c@{}}
\toprule
Method & Reference shot & Text & Exemplars \\
\midrule
VACE-14B & middle frame & \cmark & \cmark \\
UniVideo & middle frame & \cmark & \cmark \\
Kiwi-Edit & middle frame & \cmark & \xmark \\
MiniMax-H3 & video & \cmark & \cmark \\
Kling 3.0 Omni & 3 frames & \cmark & \cmark \\
Wan 3.0 & video & \cmark & \cmark \\
\rowcolor{oursrow}
\textbf{\ours (ours)} & video & \cmark & \cmark \\
\bottomrule
\end{tabularx}
\end{table}

\section{Additional Results}
\label{supp:results}
\begin{figure*}[p]
\centering
\begin{animateinline}[poster=first,width=\textwidth]{100}
\includegraphics[width=\textwidth]{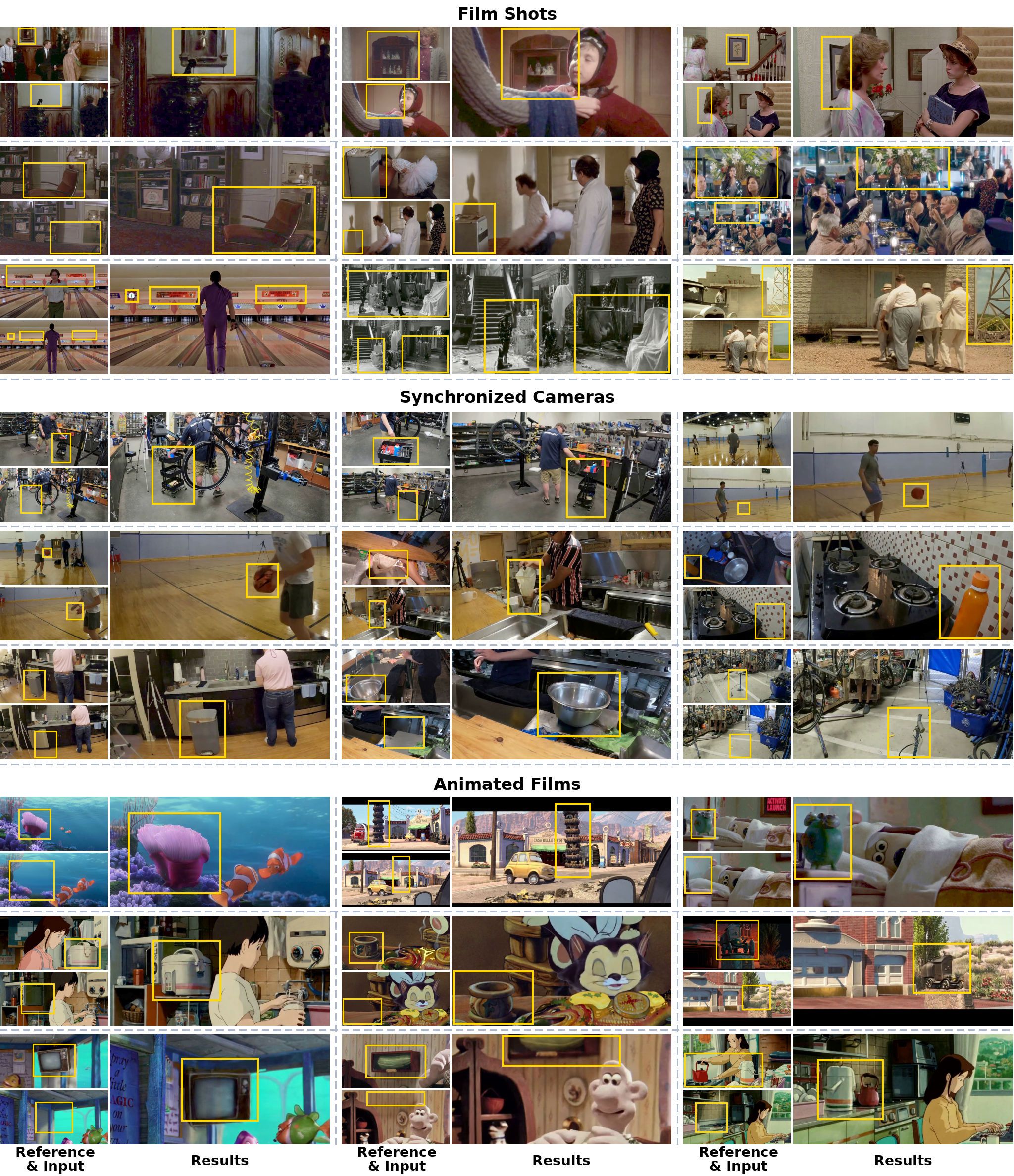}
\newframe[2]
\multiframe{8}{i=0+1}{\includegraphics[width=\textwidth]{figures/supp_results_all/f_\i.jpg}}
\end{animateinline}
\caption{Results of \ours on film shots, synchronized cameras and animated films. Yellow boxes mark the object in the reference video, the input video and the result. Readers can click and play the video clips in this figure using \textcolor{red}{\textbf{Adobe Acrobat}}.}
\label{fig:supp_results}
\end{figure*}

\mbox{\cref{fig:supp_results}} shows further results. On film shots, the reference and the target often differ strongly in scale, and the inserted objects keep their place behind the actors. On synchronized cameras, the inserted object follows the motion seen in the reference, such as a strainer held in the hand and a bouncing basketball. On animated films, the same model handles computer-animated, hand-drawn and stop-motion films.

\begin{figure*}[p]
\centering
\begin{animateinline}[poster=first,width=\textwidth]{100}
\includegraphics[width=\textwidth]{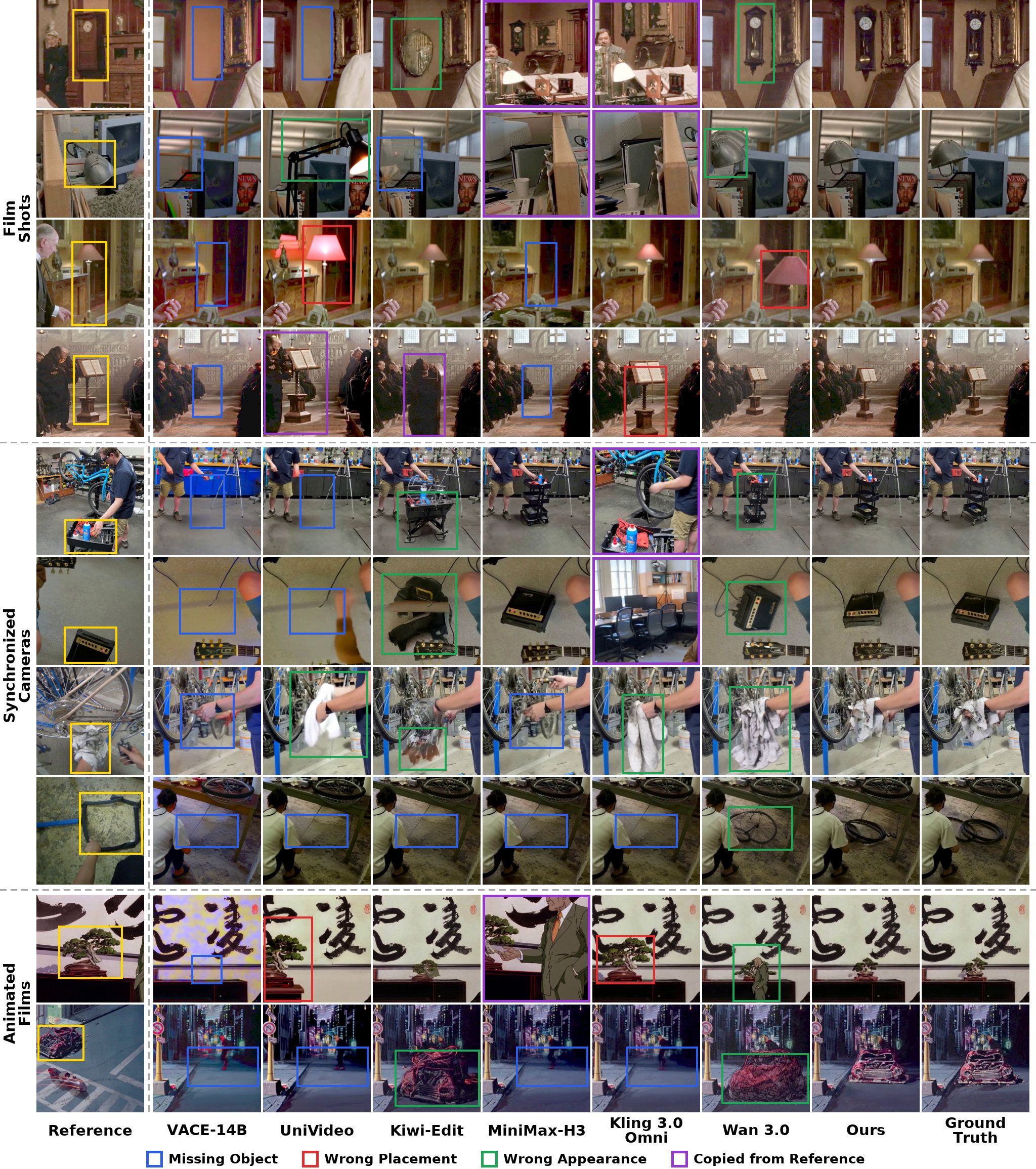}
\newframe[2]
\multiframe{8}{i=0+1}{\includegraphics[width=\textwidth]{figures/supp_comparison_big/f_\i.jpg}}
\end{animateinline}
\caption{More comparisons with the six video editors on films, synchronized cameras and animated films, with the failures marked as in \cref{fig:qualitative}; yellow boxes, shown before playback, mark the object in the reference. Readers can click and play the video clips in this figure using \textcolor{red}{\textbf{Adobe Acrobat}}.}
\label{fig:supp_comparison}
\end{figure*}

In the comparison on ten further clips (\cref{fig:supp_comparison}), the baselines leave the object out, draw a different object, show another view of the scene or misplace it, while \ours inserts each object where the ground truth shows it.

\begin{figure*}[t]
\centering
\begin{animateinline}[poster=first,width=\textwidth]{100}
\includegraphics[width=\textwidth]{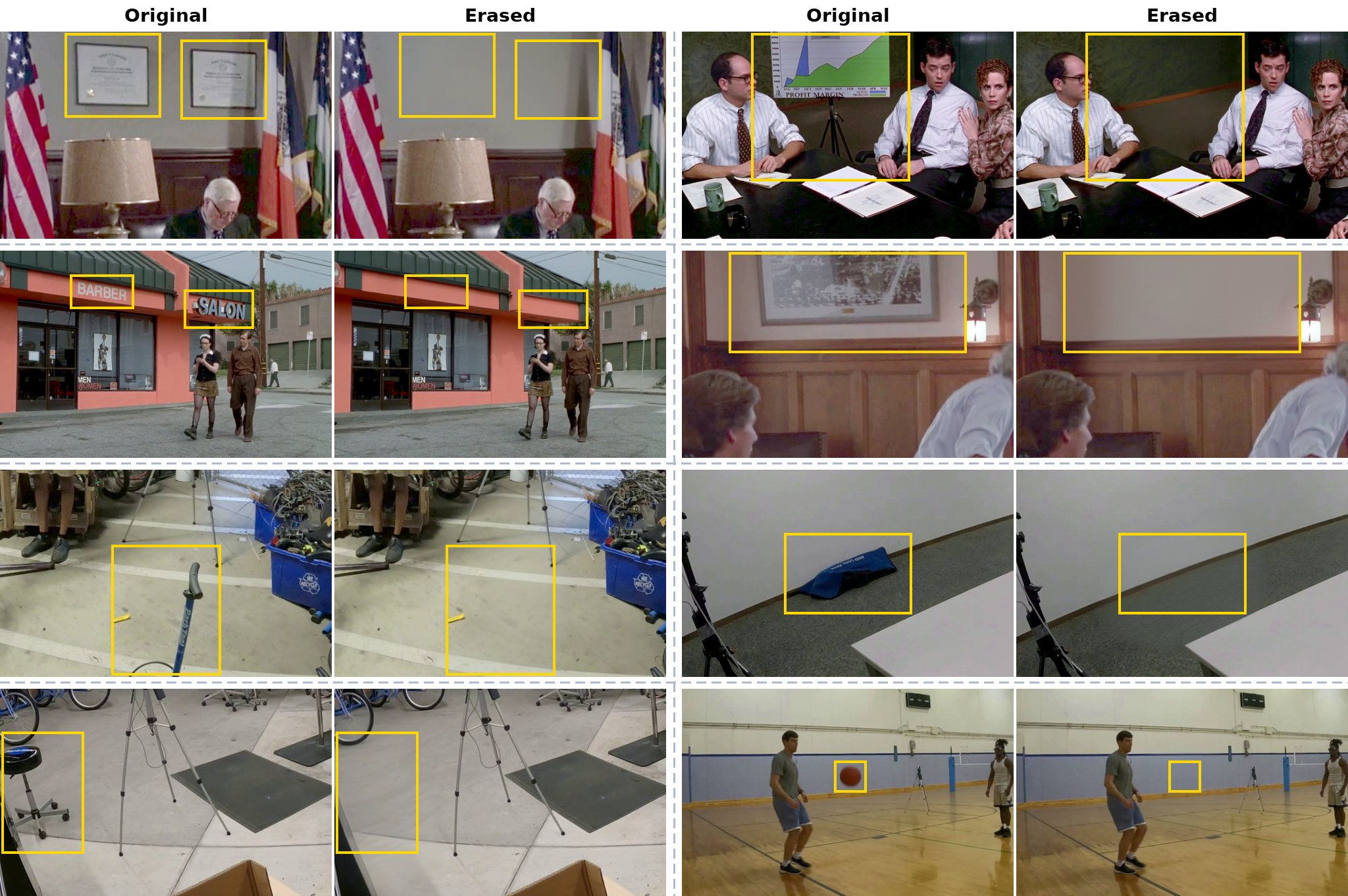}
\newframe[2]
\multiframe{8}{i=0+1}{\includegraphics[width=\textwidth]{figures/supp_erase/f_\i.jpg}}
\end{animateinline}
\caption{Examples from the data engine: the original shot (left) and the erased input (right); yellow boxes mark the removed object. Readers can click and play the video clips in this figure using \textcolor{red}{\textbf{Adobe Acrobat}}.}
\label{fig:supp_erase}
\end{figure*}

\section{Data Engine Details}
\label{supp:data}
The training set contains 17,840 clips (examples in \cref{fig:supp_erase}): 7,994 come from 769 feature films and 9,846 come from 2,310 Ego-Exo4D recordings made at 12 sites.
Of the Ego-Exo4D clips, 8,265 pair two fixed exocentric cameras and 1,581 pair the head-mounted egocentric camera with a fixed camera.
\mbox{\cref{tab:supp_stats}} gives the statistics of the training set and the benchmark.

\begin{table}[t]
\centering
\caption{Statistics of the training set and the benchmark. All clips are $1280\times704$ pixels.}
\label{tab:supp_stats}
\footnotesize
\setlength{\tabcolsep}{4pt}
\renewcommand{\arraystretch}{1.1}
\begin{tabularx}{\linewidth}{@{}X rr@{}}
\toprule
 & Training & Benchmark \\
\midrule
Clips & 17,840 & 100 \\
\quad from feature films & 7,994 & 60 \\
\quad from Ego-Exo4D & 9,846 & 40 \\
Feature films & 769 & 57 \\
Ego-Exo4D recordings & 2,310 & 40 \\
Ego-Exo4D activities & 6 & 6 \\
Frames per clip & 49, 65, 81 & 49, 65, 81 \\
Clips with two or more objects (\%) & 6.10 & 6.00 \\
\bottomrule
\end{tabularx}
\end{table}
For film shots, a vision-language model, Gemini~3.8 Flash~\cite{gemini}, watches each candidate pair of shots of one scene and lists up to five objects that can be erased from one shot and restored from the other (\cref{fig:supp_vlm_prompt}).
For each object, it gives the name, the parts, the shot to erase it from, a time and a bounding box.

When building the data, the engine also randomly erases a few decoy regions besides the object. Each decoy region is close to the object in size, between 0.7 and 1.3 times its bounding box, and lies at a plausible place elsewhere in the shot. The ground truth used for training contains the same erased decoys, so it differs from the input video only by the object.

We checked 200 clips from the data engine by eye, and in almost all of them the object is removed completely; \mbox{\cref{fig:supp_erase}} shows examples.

We also ran another test: in the input videos of 23 benchmark clips, we erase one more object of similar size and shape with the same data engine, keeping the reference video, the exemplars and the seed unchanged.
The results stay almost the same: the placement IoU changes from 0.687 to 0.683, and CLIP-I, DINO-I and LPIPS change by less than 0.002.

\section{Geometry-Aware Attention Details}
\label{supp:geometry}
\paragraph{Matching points.}
For a target token $u$ and the cells $v$ of one reference frame, the correspondence head gives $p_u(v)=\mathrm{softmax}_v A(u,v)$ (\cref{eq:corr}).
With $x_v$ denoting the image position of $v$, the matching point of $u$ and its spread are
\begin{equation}
m_u=\sum_v p_u(v)\,x_v,\qquad s_u=\sum_v p_u(v)\,\lVert x_v-m_u\rVert^2 .
\label{eq:supp_match}
\end{equation}

\begin{figure}[t]
\centering
\fbox{\begin{minipage}{0.95\linewidth}
\scriptsize\setlength{\parskip}{3pt}
These are two shots of the same film scene: shot A (the first video) and shot B (the second video).

We erase a scene object (anything except the people) from one shot, and a model learns to put it back exactly as it was by looking at the other shot [...]. Either shot can serve as the reference, chosen per object. [...]

The one thing that matters: for an object erased from one shot, does the other shot give enough information, what the object looks like and where it sits in the scene, to restore it faithfully? And can it be erased cleanly [...]? [...] If the two shots are taken from (almost) the same camera position and differ mainly by zoom, judge the pair in the same way but mark it with ``same\_view'': true.

What to erase must leave a scene that looks complete and physically possible, and must be worth learning:\\
- Erase an object together with everything resting on it or inside it [...].\\
- If several similar items form a group (a pile, a row, a stack), the group is erased as a whole. Do not pick one of several look-alike objects that the other shot cannot tell apart.\\
- Prefer large, clearly visible objects [...] over small props.\\
- The other shot must show the object clearly, not too dark, blurred or tiny, and it must look the same in both shots [...].

Answer with one JSON object only: \{``usable'', ``same\_view'', ``reason'', ``objects'': [\{``name'', ``parts'', ``erase\_from'', ``time'', ``box''\}]\}. List the erasable objects best first, at most 5.
\end{minipage}}
\caption{Instruction given to the vision-language model for a pair of film shots, abridged; [...] marks omitted text.}
\label{fig:supp_vlm_prompt}
\end{figure}

\paragraph{Fitting $F$.}
Before denoising, \ours runs its first blocks once on the input video with a little noise, together with the reference, the exemplars and the text, and computes $m_u$ and $s_u$ for probe tokens spread evenly over the target video.
With $x_u$ denoting the position of $u$ and $\tilde{x}$ denoting the homogeneous coordinates of $x$, $F$ is the weighted eight-point solution~\cite{hartley8pt}
\begin{equation}
F=\operatorname*{arg\,min}_{\lVert F\rVert=1}\;\sum_u w_u\big(\tilde{x}_u^{\top}F\,\tilde{m}_u\big)^2 ,
\label{eq:supp_fit}
\end{equation}
which is solved again in each reweighting round with
\begin{equation}
w_u=\frac{1}{s_u+\kappa}\cdot\frac{1}{1+r_u^2/\delta^2},
\label{eq:supp_w}
\end{equation}
where $\kappa$ is a small constant, $r_u$ is the Sampson distance of $(x_u,m_u)$ to the current $F$ ($r_u=0$ before the first fit), and the scale $\delta$ shrinks from round to round.
Uncertain points and points far from their epipolar lines thus lose weight.
$F$ is fitted over short time segments, over which camera motion is usually small, so that one $F$ describes most points.

\paragraph{Confidence.}
From the same points, \ours computes a confidence
\begin{equation}
c=g_r(\bar{r})\,g_s(\bar{s}),
\label{eq:supp_conf}
\end{equation}
where $\bar{r}$ is the weighted mean Sampson distance of the points to $F$, $\bar{s}$ is their mean spread, and $g_r$ and $g_s$ decrease linearly from one to zero and are clipped to $[0,1]$.
When $c$ is zero, the epipolar bias vanishes, and the attention to the reference relies on the model's own matching.
$F$ and $c$ stay fixed for all denoising steps.

\paragraph{Epipolar attention bias.}
In \cref{eq:epibias}, $d(u,v)=|\hat{\ell}_v^{\top}\tilde{x}_u|$, where $\hat{\ell}_v$ is the epipolar line $F\tilde{x}_v$ scaled so that its first two entries have unit norm, and
\begin{equation}
Z_u=\frac{N}{\sum_{v'}\exp\big(-\alpha_{b,h}\,c\,d(u,v')^{2}\big)},
\label{eq:supp_z}
\end{equation}
where $v'$ runs over the $N$ reference tokens of the same time segment as $u$, the only reference tokens that receive the bias.

\paragraph{Exemplar routing.}
DINOv2 features of the object patches of exemplar $n$ are compared with those of every reference cell, and the peak of the resulting similarity map, found by mean shift and tracked from frame to frame, gives $o_n$.
For target frame $\tau$, $F_\tau$ is fitted with RANSAC~\cite{ransac} to the most likely matching cells, $\arg\max_v A(u,v)$, of all tokens $u$ of frame $\tau$, so that the line $\ell_{n,\tau}$ follows the target camera from frame to frame.
In \cref{eq:route}, $\bar{d}_n(u)=\max\big(0,\,d(u,\ell_{n,\tau})-\Delta\big)$ with a band half-width $\Delta$, $Z_{n,\tau}$ plays the role of $Z_u$ over the tokens of frame $\tau$, and
\begin{equation}
c_{n,\tau}=g_\nu(\nu_\tau)\,g_h(h_n),
\label{eq:supp_route_conf}
\end{equation}
where $\nu_\tau$ is the weighted inlier fraction of the RANSAC fit, $h_n$ is the mass of the similarity map around $o_n$, and $g_\nu$ and $g_h$ increase linearly from zero to one and are clipped to $[0,1]$.

\begin{figure*}[t]
\centering
\begin{animateinline}[poster=first,width=\textwidth]{100}
\includegraphics[width=\textwidth]{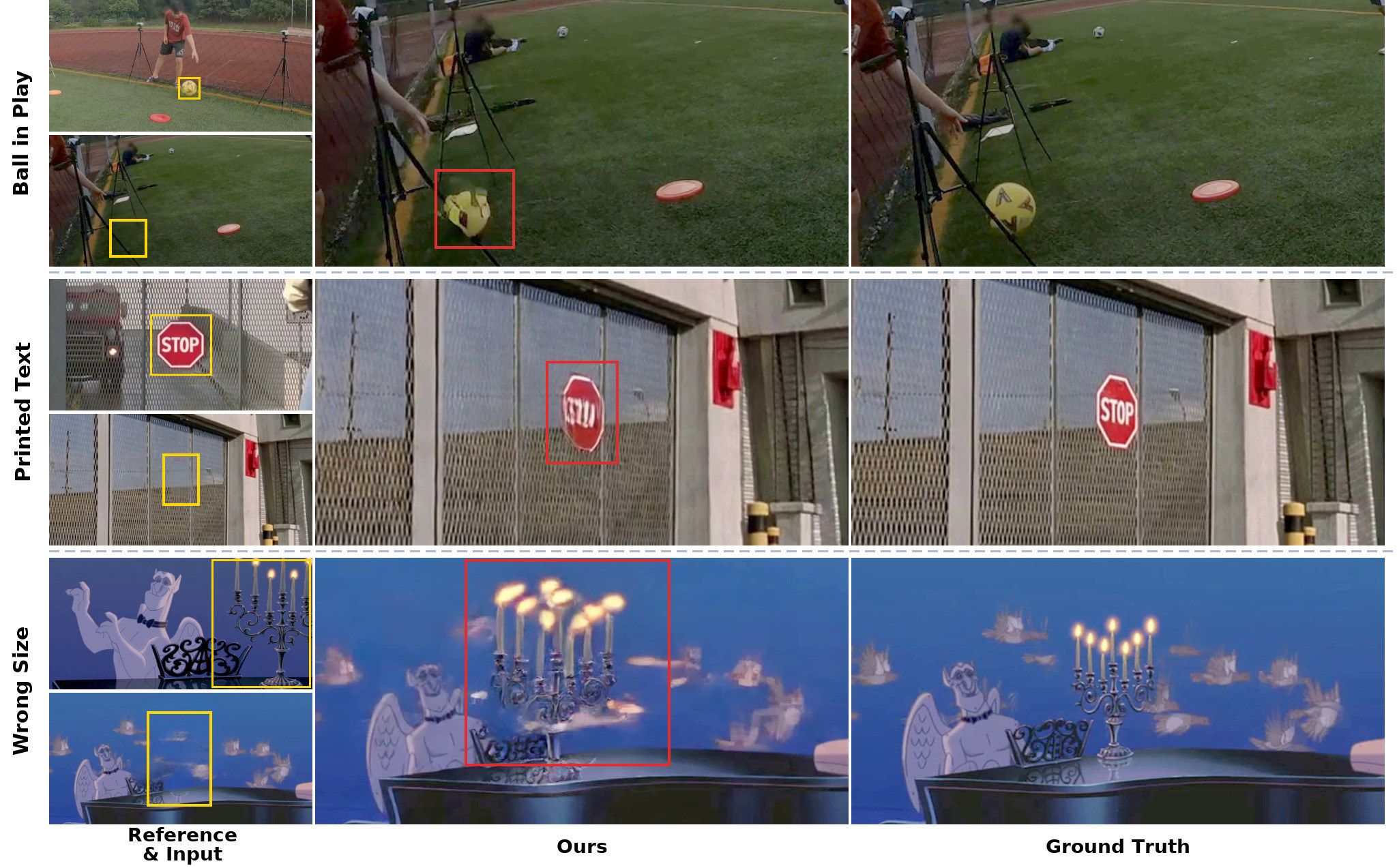}
\newframe[2]
\multiframe{8}{i=0+1}{\includegraphics[width=\textwidth]{figures/supp_failures/f_\i.jpg}}
\end{animateinline}
\caption{Failure cases of \ours (red boxes). Top: a ball in play is placed where the ground truth shows it, but it is smeared and deformed. Middle: the letters of a stop sign are not legible, and the sign is tilted. Bottom: a candelabra is placed correctly but drawn larger than in the target shot, with oversized flames. Readers can click and play the video clips in this figure using \textcolor{red}{\textbf{Adobe Acrobat}}.}
\label{fig:supp_failures}
\end{figure*}

\section{Efficiency and Inference Cost}
\label{supp:cost}
\begin{table}[!h]
\centering
\caption{Generation time per clip and training time per step on one RTX 5090 for growing reference lengths (81-frame target), with the pooled reference of \ours and with the full reference (the same backbone without pooling and without the geometry-aware attention). OOM: out of memory.}
\label{tab:supp_speed}
\footnotesize
\setlength{\tabcolsep}{4pt}
\renewcommand{\arraystretch}{1.15}
\begin{tabularx}{\linewidth}{@{}l*{4}{>{\centering\arraybackslash}X}@{}}
\toprule
Reference frames & 33 & 65 & 97 & 129 \\
\midrule
Generation, full reference (s) & 232 & 309 & 411 & 506 \\
\rowcolor{oursrow}
Generation, \ours (s) & \textbf{179} & \textbf{192} & \textbf{211} & \textbf{225} \\
Training step, full reference (s) & 14.6 & OOM & OOM & OOM \\
\rowcolor{oursrow}
Training step, \ours (s) & \textbf{12.9} & \textbf{14.1} & \textbf{15.6} & \textbf{16.6} \\
Training memory, \ours (GB) & 26.3 & 27.3 & 28.4 & 29.5 \\
\bottomrule
\end{tabularx}
\end{table}
\cref{tab:supp_speed} shows that pooling the reference makes \ours faster than the same backbone with the full reference, although the geometry-aware attention is added, and that the saving grows with the length of the reference.
From 33 to 129 reference frames, the generation time of \ours rises by about a quarter, from 179\,s to 225\,s, whereas with the full reference it more than doubles, from 232\,s to 506\,s.
\ours thus generates a clip 1.30 times faster with 33 reference frames and 2.25 times faster with 129 reference frames.
Training shows the same trend: a training step of \ours takes 12.9\,s with 33 reference frames and 16.6\,s with 129 reference frames, while training with the full reference no longer fits into the 32\,GB of one GPU beyond 33 frames.
Pooling therefore lets \ours read longer reference shots at little extra cost, both in training and at inference.

\cref{tab:cost} compares the cost of editing one benchmark clip; for \ours, the measured generation time on one RTX 5090 is priced at \$0.556 per hour.
\ours costs \$0.0258 per clip, 11.6 times less than MiniMax-H3, the least expensive commercial editor, and 28.4 times less than Wan~3.0.

\section{Failure Cases, Limitations and Future Work}
\label{supp:failures}

While \ours places objects reliably across films and synchronized cameras, it still has limitations, illustrated in \cref{fig:supp_failures}.
First, small objects in fast motion lose their shape: the ball in the first row stands where the ground truth shows it but is deformed, and the seven clips with a ball in play have the lowest DINO-I scores of the benchmark.
Second, fine printed text, such as the letters of the stop sign in the second row, is not reproduced, which remains difficult for reference-based generation in general~\cite{fineforge}.
Third, the size of the object can be wrong when the reference shot shows the scene from much closer than the target shot, as for the candelabra in the third row.

Addressing these limitations, together with handling shots longer than the 81-frame windows of \ours, removing objects across shots and editing several shots of a scene at once, remains an important direction for our future work.

\section{Ethics Statement}
\label{supp:ethics}

\ours is designed for film and video post-production, for example to keep an object consistent across the shots of a scene. Like other video editing methods, it could also be misused to add objects to real footage and mislead viewers. Edited videos should be clearly labeled as edited, and invisible watermarks and provenance metadata on generated content can help to identify such edits. The feature films and the Ego-Exo4D recordings are used for research only, and Ego-Exo4D is used under its license agreement.

In the user study, participation was voluntary. Before starting, every participant agreed to a consent statement: the answers are anonymous, only the choices, the time spent on each screen and basic browser information are recorded, the answers may be used in a research publication, and participants can stop at any time. The study was not reviewed by an Institutional Review Board, since it was an anonymous, minimal-risk rating of generated videos that recorded no names or contact details.

%% file: main.bbl
\begin{thebibliography}{102}
\providecommand{\natexlab}[1]{#1}
\providecommand{\url}[1]{\texttt{#1}}
\expandafter\ifx\csname urlstyle\endcsname\relax
  \providecommand{\doi}[1]{doi: #1}\else
  \providecommand{\doi}{doi: \begingroup \urlstyle{rm}\Url}\fi

\bibitem[{Alibaba Cloud}(2026)]{wan3}
{Alibaba Cloud}.
\newblock {Wan3.0 Video Generation}, 2026.
\newblock Alibaba Cloud Model Studio documentation, accessed 2026-10-22.

\bibitem[An et~al.(2026)An, Jia, Qiu, Zhou, Huang, Liu, Ren, Kahatapitiya, Liu,
  He, Zhang, Xiang, Yang, Belongie, and Xie]{onestory}
Zhaochong An, Menglin Jia, Haonan Qiu, Zijian Zhou, Xiaoke Huang, Zhiheng Liu,
  Weiming Ren, Kumara Kahatapitiya, Ding Liu, Sen He, Chenyang Zhang, Tao
  Xiang, Fanny Yang, Serge Belongie, and Tian Xie.
\newblock {OneStory: Coherent Multi-Shot Video Generation with Adaptive
  Memory}.
\newblock In \emph{Proceedings of the IEEE/CVF Conference on Computer Vision
  and Pattern Recognition (CVPR)}, pages 16173--16184, 2026.

\bibitem[Bai et~al.(2025{\natexlab{a}})Bai, Xia, Fu, Wang, Mu, Cao, Liu, Hu,
  Bai, Wan, and Zhang]{recammaster}
Jianhong Bai, Menghan Xia, Xiao Fu, Xintao Wang, Lianrui Mu, Jinwen Cao, Zuozhu
  Liu, Haoji Hu, Xiang Bai, Pengfei Wan, and Di Zhang.
\newblock {ReCamMaster: Camera-Controlled Generative Rendering from A Single
  Video}.
\newblock In \emph{Proceedings of the IEEE/CVF International Conference on
  Computer Vision (ICCV)}, pages 14834--14844, 2025{\natexlab{a}}.

\bibitem[Bai et~al.(2025{\natexlab{b}})Bai, Xia, Wang, Yuan, Liu, Hu, Wan, and
  Zhang]{syncammaster}
Jianhong Bai, Menghan Xia, Xintao Wang, Ziyang Yuan, Zuozhu Liu, Haoji Hu,
  Pengfei Wan, and Di Zhang.
\newblock {SynCamMaster: Synchronizing Multi-Camera Video Generation from
  Diverse Viewpoints}.
\newblock In \emph{International Conference on Learning Representations
  (ICLR)}, 2025{\natexlab{b}}.

\bibitem[Barath et~al.(2020)Barath, Noskova, Ivashechkin, and Matas]{magsacpp}
Daniel Barath, Jana Noskova, Maksym Ivashechkin, and Jiri Matas.
\newblock {MAGSAC++, a Fast, Reliable and Accurate Robust Estimator}.
\newblock In \emph{Proceedings of the IEEE/CVF Conference on Computer Vision
  and Pattern Recognition (CVPR)}, pages 1301--1309, 2020.

\bibitem[Bhalgat et~al.(2023)Bhalgat, Henriques, and Zisserman]{lighttouch}
Yash Bhalgat, Jo{\~a}o~F. Henriques, and Andrew Zisserman.
\newblock {A Light Touch Approach to Teaching Transformers Multi-View
  Geometry}.
\newblock In \emph{Proceedings of the IEEE/CVF Conference on Computer Vision
  and Pattern Recognition (CVPR)}, pages 4958--4969, 2023.

\bibitem[Bian et~al.(2026)Bian, Chen, Li, Zhi, Sang, Luo, and Xu]{vap}
Yuxuan Bian, Xin Chen, Zenan Li, Tiancheng Zhi, Shen Sang, Linjie Luo, and
  Qiang Xu.
\newblock {Video-As-Prompt: Unified Semantic Control for Video Generation}.
\newblock In \emph{International Conference on Learning Representations
  (ICLR)}, 2026.

\bibitem[Blattmann et~al.(2023{\natexlab{a}})Blattmann, Dockhorn, Kulal,
  Mendelevitch, Kilian, Lorenz, Levi, English, Voleti, Letts, Jampani, and
  Rombach]{svd}
Andreas Blattmann, Tim Dockhorn, Sumith Kulal, Daniel Mendelevitch, Maciej
  Kilian, Dominik Lorenz, Yam Levi, Zion English, Vikram Voleti, Adam Letts,
  Varun Jampani, and Robin Rombach.
\newblock {Stable Video Diffusion: Scaling Latent Video Diffusion Models to
  Large Datasets}.
\newblock \emph{arXiv preprint arXiv:2311.15127}, 2023{\natexlab{a}}.

\bibitem[Blattmann et~al.(2023{\natexlab{b}})Blattmann, Rombach, Ling,
  Dockhorn, Kim, Fidler, and Kreis]{videoldm}
Andreas Blattmann, Robin Rombach, Huan Ling, Tim Dockhorn, Seung~Wook Kim,
  Sanja Fidler, and Karsten Kreis.
\newblock {Align your Latents: High-Resolution Video Synthesis with Latent
  Diffusion Models}.
\newblock In \emph{Proceedings of the IEEE/CVF Conference on Computer Vision
  and Pattern Recognition (CVPR)}, pages 22563--22575, 2023{\natexlab{b}}.

\bibitem[Cai et~al.(2026)Cai, Yang, Zhang, Guo, Xiao, Yang, Xu, Yang, Yuille,
  Guibas, Agrawala, Jiang, and Wetzstein]{moc}
Shengqu Cai, Ceyuan Yang, Lvmin Zhang, Yuwei Guo, Junfei Xiao, Ziyan Yang,
  Yinghao Xu, Zhenheng Yang, Alan Yuille, Leonidas Guibas, Maneesh Agrawala, Lu
  Jiang, and Gordon Wetzstein.
\newblock {Mixture of Contexts for Long Video Generation}.
\newblock In \emph{International Conference on Learning Representations
  (ICLR)}, 2026.

\bibitem[Cao et~al.(2024{\natexlab{a}})Cao, Cai, Dong, Wang, and
  Fu]{leftrefill}
Chenjie Cao, Yunuo Cai, Qiaole Dong, Yikai Wang, and Yanwei Fu.
\newblock {LeftRefill: Filling Right Canvas based on Left Reference through
  Generalized Text-to-Image Diffusion Model}.
\newblock In \emph{Proceedings of the IEEE/CVF Conference on Computer Vision
  and Pattern Recognition (CVPR)}, pages 7705--7715, 2024{\natexlab{a}}.

\bibitem[Cao et~al.(2024{\natexlab{b}})Cao, Yu, Wang, Xue, and Fu]{mvinpainter}
Chenjie Cao, Chaohui Yu, Fan Wang, Xiangyang Xue, and Yanwei Fu.
\newblock {MVInpainter: Learning Multi-View Consistent Inpainting to Bridge 2D
  and 3D Editing}.
\newblock In \emph{Advances in Neural Information Processing Systems
  (NeurIPS)}, pages 99299--99332, 2024{\natexlab{b}}.

\bibitem[Caron et~al.(2021)Caron, Touvron, Misra, J{\'e}gou, Mairal,
  Bojanowski, and Joulin]{dino}
Mathilde Caron, Hugo Touvron, Ishan Misra, Herv{\'e} J{\'e}gou, Julien Mairal,
  Piotr Bojanowski, and Armand Joulin.
\newblock {Emerging Properties in Self-Supervised Vision Transformers}.
\newblock In \emph{Proceedings of the IEEE/CVF International Conference on
  Computer Vision (ICCV)}, pages 9650--9660, 2021.

\bibitem[Chang et~al.(2023)Chang, Yu, and Zhang]{sem}
Jiahao Chang, Jiahuan Yu, and Tianzhu Zhang.
\newblock {Structured Epipolar Matcher for Local Feature Matching}.
\newblock In \emph{Proceedings of the IEEE/CVF Conference on Computer Vision
  and Pattern Recognition (CVPR) Workshops}, pages 6177--6186, 2023.

\bibitem[Charatan et~al.(2024)Charatan, Li, Tagliasacchi, and
  Sitzmann]{pixelsplat}
David Charatan, Sizhe~Lester Li, Andrea Tagliasacchi, and Vincent Sitzmann.
\newblock {pixelSplat: 3D Gaussian Splats from Image Pairs for Scalable
  Generalizable 3D Reconstruction}.
\newblock In \emph{Proceedings of the IEEE/CVF Conference on Computer Vision
  and Pattern Recognition (CVPR)}, pages 19457--19467, 2024.

\bibitem[Chen et~al.(2025)Chen, Wu, Chen, Teng, Gao, Feng, Qin, and Zhao]{gafm}
Gonglin Chen, Jinsen Wu, Haiwei Chen, Wenbin Teng, Zhiyuan Gao, Andrew Feng,
  Rongjun Qin, and Yajie Zhao.
\newblock {Geometry-Aware Feature Matching for Large-Scale Structure from
  Motion}.
\newblock In \emph{International Conference on 3D Vision (3DV)}, pages 34--43,
  2025.

\bibitem[Chen et~al.(2026{\natexlab{a}})Chen, Li, Bai, Ma, Zhang, Li, Chen, Li,
  Liu, Zhao, Li, and He]{omniinsert}
Jinshu Chen, Xinghui Li, Xu Bai, Tianxiang Ma, Pengze Zhang, Mengtian Li,
  Zhuowei Chen, Gen Li, Lijie Liu, Songtao Zhao, Bingchuan Li, and Qian He.
\newblock {OmniInsert: Mask-Free Video Insertion of Any Reference via Diffusion
  Transformer Models}.
\newblock In \emph{Proceedings of the IEEE/CVF Conference on Computer Vision
  and Pattern Recognition (CVPR) Findings}, pages 4067--4077,
  2026{\natexlab{a}}.

\bibitem[Chen et~al.(2024)Chen, Huang, Liu, Shen, Zhao, and Zhao]{anydoor}
Xi Chen, Lianghua Huang, Yu Liu, Yujun Shen, Deli Zhao, and Hengshuang Zhao.
\newblock {AnyDoor: Zero-shot Object-level Image Customization}.
\newblock In \emph{Proceedings of the IEEE/CVF Conference on Computer Vision
  and Pattern Recognition (CVPR)}, pages 6593--6602, 2024.

\bibitem[Chen et~al.(2026{\natexlab{b}})Chen, Chu, Gleize, Liang, Sax, Tang,
  Wang, Guo, Hardin, Li, et~al.]{sam3d}
Xingyu Chen, Fu-Jen Chu, Pierre Gleize, Kevin~J. Liang, Alexander Sax, Hao
  Tang, Weiyao Wang, Michelle Guo, Thibaut Hardin, Xiang Li, et~al.
\newblock {SAM 3D: 3Dfy Anything in Images}.
\newblock In \emph{Proceedings of the IEEE/CVF Conference on Computer Vision
  and Pattern Recognition (CVPR)}, pages 7220--7232, 2026{\natexlab{b}}.

\bibitem[Esser et~al.(2021)Esser, Rombach, and Ommer]{vqgan}
Patrick Esser, Robin Rombach, and Bj{\"o}rn Ommer.
\newblock {Taming Transformers for High-Resolution Image Synthesis}.
\newblock In \emph{Proceedings of the IEEE/CVF Conference on Computer Vision
  and Pattern Recognition (CVPR)}, pages 12873--12883, 2021.

\bibitem[Esser et~al.(2023)Esser, Chiu, Atighehchian, Granskog, and
  Germanidis]{gen1}
Patrick Esser, Johnathan Chiu, Parmida Atighehchian, Jonathan Granskog, and
  Anastasis Germanidis.
\newblock {Structure and Content-Guided Video Synthesis with Diffusion Models}.
\newblock In \emph{Proceedings of the IEEE/CVF International Conference on
  Computer Vision (ICCV)}, pages 7346--7356, 2023.

\bibitem[Feng et~al.(2026)Feng, Ma, Wang, Wang, Qin, Cheng, Li, Chen, and
  Wang]{mseditor}
Kunyu Feng, Yue Ma, Bingyuan Wang, Yuefeng Wang, Zhiyuan Qin, Hao Cheng, Hao
  Li, Qifeng Chen, and Zeyu Wang.
\newblock {MSEditor: Toward Consistent Multi-Shot Video Editing}.
\newblock In \emph{Proceedings of the European Conference on Computer Vision
  (ECCV)}, pages 76--95, 2026.

\bibitem[Fischler and Bolles(1981)]{ransac}
Martin~A. Fischler and Robert~C. Bolles.
\newblock {Random Sample Consensus: A Paradigm for Model Fitting with
  Applications to Image Analysis and Automated Cartography}.
\newblock \emph{Communications of the ACM}, 24\penalty0 (6):\penalty0 381--395,
  1981.

\bibitem[Gao et~al.(2026{\natexlab{a}})Gao, Tang, and Gao]{fineforge}
Heyuan Gao, Bangxun Tang, and Zeyu Gao.
\newblock {FineForge: Dual-Driven Reinforcement Learning for Fine-Grained
  Detail Generation in Diffusion Models}.
\newblock In \emph{Proceedings of the ACM International Conference on
  Multimedia (MM)}, pages 2178--2186, 2026{\natexlab{a}}.

\bibitem[Gao et~al.(2026{\natexlab{b}})Gao, Tang, Song, Fang, He, Yang, and
  Shou]{paistudio}
Heyuan Gao, Bangxun Tang, Yiren Song, Guian Fang, Zijian He, Jie Yang, and
  Mike~Zheng Shou.
\newblock {PAI-Studio: Cinematic Video Background Replacement with Camera-Aware
  Motion}.
\newblock In \emph{Proceedings of the European Conference on Computer Vision
  (ECCV)}, pages 455--472, 2026{\natexlab{b}}.

\bibitem[{Gemini Team} et~al.(2023){Gemini Team}, Anil, Borgeaud, Alayrac, Yu,
  Soricut, Schalkwyk, Dai, Hauth, Millican, et~al.]{gemini}
{Gemini Team}, Rohan Anil, Sebastian Borgeaud, Jean-Baptiste Alayrac, Jiahui
  Yu, Radu Soricut, Johan Schalkwyk, Andrew~M. Dai, Anja Hauth, Katie Millican,
  et~al.
\newblock {Gemini: A Family of Highly Capable Multimodal Models}.
\newblock \emph{arXiv preprint arXiv:2312.11805}, 2023.

\bibitem[Geyer et~al.(2024)Geyer, Bar-Tal, Bagon, and Dekel]{tokenflow}
Michal Geyer, Omer Bar-Tal, Shai Bagon, and Tali Dekel.
\newblock {TokenFlow: Consistent Diffusion Features for Consistent Video
  Editing}.
\newblock In \emph{International Conference on Learning Representations
  (ICLR)}, 2024.

\bibitem[Grauman et~al.(2024)Grauman, Westbury, Torresani, Kitani, Malik,
  Afouras, Ashutosh, Baiyya, Bansal, Boote, et~al.]{egoexo4d}
Kristen Grauman, Andrew Westbury, Lorenzo Torresani, Kris Kitani, Jitendra
  Malik, Triantafyllos Afouras, Kumar Ashutosh, Vijay Baiyya, Siddhant Bansal,
  Bikram Boote, et~al.
\newblock {Ego-Exo4D: Understanding Skilled Human Activity from First- and
  Third-Person Perspectives}.
\newblock In \emph{Proceedings of the IEEE/CVF Conference on Computer Vision
  and Pattern Recognition (CVPR)}, pages 19383--19400, 2024.

\bibitem[Guo et~al.(2026)Guo, Wei, Li, Huang, Liu, Luo, Wei, Zhu, Wang, Xu,
  Wang, Wan, Fan, and Hou]{contextmaster}
Xu Guo, Zhengxuan Wei, Xinghui Li, Hanzhuo Huang, Xinyu Liu, Xiangyang Luo, Min
  Wei, Yiran Zhu, Qiulin Wang, Yulong Xu, Xintao Wang, Pengfei Wan, Qi Fan, and
  Xiangwang Hou.
\newblock {ContextMaster: Interactive Multi-Shot Video Creation via
  Fixed-Budget Sparse Context Routing}.
\newblock \emph{arXiv preprint arXiv:2608.04956}, 2026.

\bibitem[Guo et~al.(2025)Guo, Yang, Yang, Ma, Lin, Yang, Lin, and Jiang]{lct}
Yuwei Guo, Ceyuan Yang, Ziyan Yang, Zhibei Ma, Zhijie Lin, Zhenheng Yang, Dahua
  Lin, and Lu Jiang.
\newblock {Long Context Tuning for Video Generation}.
\newblock In \emph{Proceedings of the IEEE/CVF International Conference on
  Computer Vision (ICCV)}, pages 17281--17291, 2025.

\bibitem[Hartley(1997)]{hartley8pt}
Richard~I. Hartley.
\newblock {In Defense of the Eight-Point Algorithm}.
\newblock \emph{IEEE Transactions on Pattern Analysis and Machine
  Intelligence}, 19\penalty0 (6):\penalty0 580--593, 1997.

\bibitem[He et~al.(2020)He, Yan, Fragkiadaki, and Yu]{epipolartransformers}
Yihui He, Rui Yan, Katerina Fragkiadaki, and Shoou-I Yu.
\newblock {Epipolar Transformers}.
\newblock In \emph{Proceedings of the IEEE/CVF Conference on Computer Vision
  and Pattern Recognition (CVPR)}, pages 7776--7785, 2020.

\bibitem[Hu et~al.(2022)Hu, Shen, Wallis, Allen-Zhu, Li, Wang, Wang, and
  Chen]{lora}
Edward~J. Hu, Yelong Shen, Phillip Wallis, Zeyuan Allen-Zhu, Yuanzhi Li, Shean
  Wang, Lu Wang, and Weizhu Chen.
\newblock {LoRA: Low-Rank Adaptation of Large Language Models}.
\newblock In \emph{International Conference on Learning Representations
  (ICLR)}, 2022.

\bibitem[Huang et~al.(2026)Huang, Lu, Jia, Wang, Liu, and Yang]{pecontext}
Binyuan Huang, Yuning Lu, Weinan Jia, Hualiang Wang, Mu Liu, and Daiqing Yang.
\newblock {Rethinking Position Embedding as a Context Controller for
  Multi-Reference and Multi-Shot Video Generation}.
\newblock In \emph{Proceedings of the IEEE/CVF Conference on Computer Vision
  and Pattern Recognition (CVPR)}, pages 23304--23313, 2026.

\bibitem[Huang et~al.(2024{\natexlab{a}})Huang, Wang, Wu, Shi, Dou, Liang,
  Feng, Liu, and Zhou]{incontextlora}
Lianghua Huang, Wei Wang, Zhi-Fan Wu, Yupeng Shi, Huanzhang Dou, Chen Liang,
  Yutong Feng, Yu Liu, and Jingren Zhou.
\newblock {In-Context LoRA for Diffusion Transformers}.
\newblock \emph{arXiv preprint arXiv:2410.23775}, 2024{\natexlab{a}}.

\bibitem[Huang et~al.(2024{\natexlab{b}})Huang, Wen, Dong, Wang, Li, Chen, Cao,
  Liang, Qiao, Dai, and Sheng]{epidiff}
Zehuan Huang, Hao Wen, Junting Dong, Yaohui Wang, Yangguang Li, Xinyuan Chen,
  Yan-Pei Cao, Ding Liang, Yu Qiao, Bo Dai, and Lu Sheng.
\newblock {EpiDiff: Enhancing Multi-View Synthesis via Localized
  Epipolar-Constrained Diffusion}.
\newblock In \emph{Proceedings of the IEEE/CVF Conference on Computer Vision
  and Pattern Recognition (CVPR)}, pages 9784--9794, 2024{\natexlab{b}}.

\bibitem[Jeong et~al.(2025)Jeong, Huang, Ye, Mitra, and Ceylan]{track4gen}
Hyeonho Jeong, Chun-Hao~P. Huang, Jong~Chul Ye, Niloy~J. Mitra, and Duygu
  Ceylan.
\newblock {Track4Gen: Teaching Video Diffusion Models to Track Points Improves
  Video Generation}.
\newblock In \emph{Proceedings of the IEEE/CVF Conference on Computer Vision
  and Pattern Recognition (CVPR)}, pages 7276--7287, 2025.

\bibitem[Jiang et~al.(2025)Jiang, Han, Mao, Zhang, Pan, and Liu]{vace}
Zeyinzi Jiang, Zhen Han, Chaojie Mao, Jingfeng Zhang, Yulin Pan, and Yu Liu.
\newblock {VACE: All-in-One Video Creation and Editing}.
\newblock In \emph{Proceedings of the IEEE/CVF International Conference on
  Computer Vision (ICCV)}, pages 17191--17202, 2025.

\bibitem[Jin et~al.(2026)Jin, Jang, Hyung, Kim, Kim, Kim, Choi, Kim, and
  Choo]{insertanywhere}
Hoiyeong Jin, Hyojin Jang, Junha Hyung, Jeongho Kim, Kinam Kim, Dongjin Kim,
  Huijin Choi, Hyeonji Kim, and Jaegul Choo.
\newblock {InsertAnywhere: Geometrically Grounded and Optics-Aware Video Object
  Insertion}.
\newblock In \emph{Proceedings of the European Conference on Computer Vision
  (ECCV)}, pages 38--54, 2026.

\bibitem[Ju et~al.(2025)Ju, Ye, Liu, Wang, Wang, Wan, Zhang, Gai, and
  Xu]{fulldit}
Xuan Ju, Weicai Ye, Quande Liu, Qiulin Wang, Xintao Wang, Pengfei Wan, Di
  Zhang, Kun Gai, and Qiang Xu.
\newblock {FullDiT: Video Generative Foundation Models with Multimodal Control
  via Full Attention}.
\newblock In \emph{Proceedings of the IEEE/CVF International Conference on
  Computer Vision (ICCV)}, pages 15737--15747, 2025.

\bibitem[Kang et~al.(2026)Kang, Kim, Kim, Park, Hyung, and Choo]{egox}
Taewoong Kang, Kinam Kim, Dohyeon Kim, Minho Park, Junha Hyung, and Jaegul
  Choo.
\newblock {EgoX: Egocentric Video Generation from a Single Exocentric Video}.
\newblock In \emph{Proceedings of the IEEE/CVF Conference on Computer Vision
  and Pattern Recognition (CVPR)}, pages 11116--11126, 2026.

\bibitem[Kant et~al.(2024)Kant, Siarohin, Wu, Vasilkovsky, Qian, Ren, Guler,
  Ghanem, Tulyakov, and Gilitschenski]{spad}
Yash Kant, Aliaksandr Siarohin, Ziyi Wu, Michael Vasilkovsky, Guocheng Qian,
  Jian Ren, Riza~Alp Guler, Bernard Ghanem, Sergey Tulyakov, and Igor
  Gilitschenski.
\newblock {SPAD: Spatially Aware Multi-View Diffusers}.
\newblock In \emph{Proceedings of the IEEE/CVF Conference on Computer Vision
  and Pattern Recognition (CVPR)}, pages 10026--10038, 2024.

\bibitem[Kara et~al.(2025)Kara, Singh, Liu, Ceylan, Rehg, and
  Hinz]{shotadapter}
Ozgur Kara, Krishna~Kumar Singh, Feng Liu, Duygu Ceylan, James~M. Rehg, and
  Tobias Hinz.
\newblock {ShotAdapter: Text-to-Multi-Shot Video Generation with Diffusion
  Models}.
\newblock In \emph{Proceedings of the IEEE/CVF Conference on Computer Vision
  and Pattern Recognition (CVPR)}, pages 28405--28415, 2025.

\bibitem[{Kling AI}(2026)]{kling3omni}
{Kling AI}.
\newblock {Kling VIDEO 3.0 Omni Guide}, 2026.
\newblock Kling AI user guide, accessed 2026-10-22.

\bibitem[Kong et~al.(2024)Kong, Tian, Zhang, Min, Dai, Zhou, Xiong, Li, Wu,
  Zhang, et~al.]{hunyuanvideo}
Weijie Kong, Qi Tian, Zijian Zhang, Rox Min, Zuozhuo Dai, Jin Zhou, Jiangfeng
  Xiong, Xin Li, Bo Wu, Jianwei Zhang, et~al.
\newblock {HunyuanVideo: A Systematic Framework For Large Video Generative
  Models}.
\newblock \emph{arXiv preprint arXiv:2412.03603}, 2024.

\bibitem[Ku et~al.(2024)Ku, Wei, Ren, Yang, and Chen]{anyv2v}
Max Ku, Cong Wei, Weiming Ren, Huan Yang, and Wenhu Chen.
\newblock {AnyV2V: A Tuning-Free Framework For Any Video-to-Video Editing
  Tasks}.
\newblock \emph{Transactions on Machine Learning Research}, 2024.

\bibitem[Kuang et~al.(2024)Kuang, Cai, He, Xu, Li, Guibas, and Wetzstein]{cvd}
Zhengfei Kuang, Shengqu Cai, Hao He, Yinghao Xu, Hongsheng Li, Leonidas Guibas,
  and Gordon Wetzstein.
\newblock {Collaborative Video Diffusion: Consistent Multi-video Generation
  with Camera Control}.
\newblock In \emph{Advances in Neural Information Processing Systems
  (NeurIPS)}, pages 16240--16271, 2024.

\bibitem[Kwon et~al.(2026)Kwon, Choi, Park, Jeon, Jang, Seo, Kwak, Kim, and
  Kim]{cameo}
Minkyung Kwon, Jinhyeok Choi, Jiho Park, Seonghu Jeon, Jinhyuk Jang, Junyoung
  Seo, Minseop Kwak, Jin-Hwa Kim, and Seungryong Kim.
\newblock {Correspondence-Attention Alignment for Multi-View Diffusion Models}.
\newblock In \emph{Proceedings of the IEEE/CVF Conference on Computer Vision
  and Pattern Recognition (CVPR)}, pages 2316--2326, 2026.

\bibitem[Lee et~al.(2026)Lee, Jung, Lee, Kim, Kim, Narihira, Fukuda, Koo, Han,
  Mitsufuji, and Kim]{mvtrack4gen}
JoungBin Lee, Jaewoo Jung, Jongmin Lee, Tongmin Kim, Hyunsung Kim, Takuya
  Narihira, Kazumi Fukuda, Jahyeok Koo, Jisang Han, Yuki Mitsufuji, and
  Seungryong Kim.
\newblock {MVTrack4Gen: Multi-View Point Tracking as Geometric Supervision for
  4D Video Generation}.
\newblock \emph{arXiv preprint arXiv:2606.26087}, 2026.

\bibitem[Lin et~al.(2026)Lin, Liang, Zeng, Bai, Chen, and Shou]{kiwiedit}
Yiqi Lin, Guoqiang Liang, Ziyun Zeng, Zechen Bai, Yanzhe Chen, and Mike~Zheng
  Shou.
\newblock {Kiwi-Edit: Versatile Video Editing via Instruction and Reference
  Guidance}.
\newblock \emph{arXiv preprint arXiv:2603.02175}, 2026.

\bibitem[Lindenberger et~al.(2026)Lindenberger, Sarlin, Miksik, and
  Pollefeys]{flexglue}
Philipp Lindenberger, Paul-Edouard Sarlin, Ondrej Miksik, and Marc Pollefeys.
\newblock {FlexGlue: Sparse-to-Dense Feature Matching and Visual
  Disambiguation}.
\newblock \emph{arXiv preprint arXiv:2610.32976}, 2026.

\bibitem[Liu et~al.(2024)Liu, Mao, Xu, Keppo, and Shou]{exo2egov}
Jia-Wei Liu, Weijia Mao, Zhongcong Xu, Jussi Keppo, and Mike~Zheng Shou.
\newblock {Exocentric-to-Egocentric Video Generation}.
\newblock In \emph{Advances in Neural Information Processing Systems
  (NeurIPS)}, pages 136149--136172, 2024.

\bibitem[Liu et~al.(2025)Liu, Yang, Chen, Liu, and Lin]{corrfill}
Kuan-Hung Liu, Cheng-Kun Yang, Min-Hung Chen, Yu-Lun Liu, and Yen-Yu Lin.
\newblock {CorrFill: Enhancing Faithfulness in Reference-Based Inpainting with
  Correspondence Guidance in Diffusion Models}.
\newblock In \emph{Proceedings of the Winter Conference on Applications of
  Computer Vision (WACV)}, pages 1618--1627, 2025.

\bibitem[Ma et~al.(2026)Ma, Wang, Ma, Wang, Zheng, Yang, Li, Zhao, Ying, Yang,
  Liu, and Chen]{groupediting}
Yue Ma, Xinyu Wang, Qianli Ma, Qinghe Wang, Mingzhe Zheng, Xiangpeng Yang, Hao
  Li, Chongbo Zhao, Jixuan Ying, Harry Yang, Hongyu Liu, and Qifeng Chen.
\newblock {Group Editing: Edit Multiple Images in One Go}.
\newblock In \emph{Proceedings of the IEEE/CVF Conference on Computer Vision
  and Pattern Recognition (CVPR)}, pages 43418--43428, 2026.

\bibitem[Meng et~al.(2026)Meng, Ouyang, Yu, Wang, Wang, Cheng, Wang, Ma, Li,
  Chen, et~al.]{holocine}
Yihao Meng, Hao Ouyang, Yue Yu, Qiuyu Wang, Wen Wang, Ka~Leong Cheng, Hanlin
  Wang, Shuailei Ma, Yixuan Li, Cheng Chen, et~al.
\newblock {HoloCine: Holistic Generation of Cinematic Multi-Shot Long Video
  Narratives}.
\newblock In \emph{Proceedings of the IEEE/CVF Conference on Computer Vision
  and Pattern Recognition (CVPR)}, pages 461--471, 2026.

\bibitem[{MiniMax}(2026)]{minimaxh3}
{MiniMax}.
\newblock {MiniMax-H3}, 2026.
\newblock Hugging Face model card, accessed 2026-10-22.

\bibitem[Nam et~al.(2025)Nam, Son, Chung, Kim, Jin, Hur, and Kim]{difftrack}
Jisu Nam, Soowon Son, Dahyun Chung, Jiyoung Kim, Siyoon Jin, Junhwa Hur, and
  Seungryong Kim.
\newblock {Emergent Temporal Correspondences from Video Diffusion
  Transformers}.
\newblock In \emph{Advances in Neural Information Processing Systems
  (NeurIPS)}, 2025.

\bibitem[Oquab et~al.(2024)Oquab, Darcet, Moutakanni, Vo, Szafraniec, Khalidov,
  Fernandez, Haziza, Massa, El-Nouby, et~al.]{dinov2}
Maxime Oquab, Timothée Darcet, Théo Moutakanni, Huy Vo, Marc Szafraniec,
  Vasil Khalidov, Pierre Fernandez, Daniel Haziza, Francisco Massa, Alaaeldin
  El-Nouby, et~al.
\newblock {DINOv2: Learning Robust Visual Features without Supervision}.
\newblock \emph{Transactions on Machine Learning Research}, 2024.

\bibitem[Peebles and Xie(2023)]{dit}
William Peebles and Saining Xie.
\newblock {Scalable Diffusion Models with Transformers}.
\newblock In \emph{Proceedings of the IEEE/CVF International Conference on
  Computer Vision (ICCV)}, pages 4195--4205, 2023.

\bibitem[Qi et~al.(2023)Qi, Cun, Zhang, Lei, Wang, Shan, and Chen]{fatezero}
Chenyang Qi, Xiaodong Cun, Yong Zhang, Chenyang Lei, Xintao Wang, Ying Shan,
  and Qifeng Chen.
\newblock {FateZero: Fusing Attentions for Zero-shot Text-based Video Editing}.
\newblock In \emph{Proceedings of the IEEE/CVF International Conference on
  Computer Vision (ICCV)}, pages 15932--15942, 2023.

\bibitem[Radford et~al.(2021)Radford, Kim, Hallacy, Ramesh, Goh, Agarwal,
  Sastry, Askell, Mishkin, Clark, Krueger, and Sutskever]{clip}
Alec Radford, Jong~Wook Kim, Chris Hallacy, Aditya Ramesh, Gabriel Goh,
  Sandhini Agarwal, Girish Sastry, Amanda Askell, Pamela Mishkin, Jack Clark,
  Gretchen Krueger, and Ilya Sutskever.
\newblock {Learning Transferable Visual Models From Natural Language
  Supervision}.
\newblock In \emph{International Conference on Machine Learning (ICML)}, pages
  8748--8763, 2021.

\bibitem[Ranftl and Koltun(2018)]{deepfundamental}
Ren{\'e} Ranftl and Vladlen Koltun.
\newblock {Deep Fundamental Matrix Estimation}.
\newblock In \emph{Proceedings of the European Conference on Computer Vision
  (ECCV)}, pages 292--309, 2018.

\bibitem[Ravi et~al.(2025)Ravi, Gabeur, Hu, Hu, Ryali, Ma, Khedr, Rädle,
  Rolland, Gustafson, et~al.]{sam2}
Nikhila Ravi, Valentin Gabeur, Yuan-Ting Hu, Ronghang Hu, Chaitanya Ryali,
  Tengyu Ma, Haitham Khedr, Roman Rädle, Chloe Rolland, Laura Gustafson,
  et~al.
\newblock {SAM 2: Segment Anything in Images and Videos}.
\newblock In \emph{International Conference on Learning Representations
  (ICLR)}, 2025.

\bibitem[Ruiz et~al.(2023)Ruiz, Li, Jampani, Pritch, Rubinstein, and
  Aberman]{dreambooth}
Nataniel Ruiz, Yuanzhen Li, Varun Jampani, Yael Pritch, Michael Rubinstein, and
  Kfir Aberman.
\newblock {DreamBooth: Fine Tuning Text-to-Image Diffusion Models for
  Subject-Driven Generation}.
\newblock In \emph{Proceedings of the IEEE/CVF Conference on Computer Vision
  and Pattern Recognition (CVPR)}, pages 22500--22510, 2023.

\bibitem[{Runway}(2026)]{aleph2}
{Runway}.
\newblock {Introducing Aleph 2.0 and Edit Studio}, 2026.
\newblock Runway news, accessed 2026-10-04.

\bibitem[Song et~al.(2026{\natexlab{a}})Song, Song, Peng, Gao, and
  Shou]{worldwander}
Quanjian Song, Yiren Song, Kelly Peng, Yuan Gao, and Mike~Zheng Shou.
\newblock {WorldWander: Bridging Egocentric and Exocentric Worlds in Video
  Generation}.
\newblock In \emph{Proceedings of the European Conference on Computer Vision
  (ECCV)}, pages 612--630, 2026{\natexlab{a}}.

\bibitem[Song et~al.(2026{\natexlab{b}})Song, Jiang, Yang, Cheng, Quan, and
  Yang]{insertanything}
Wensong Song, Hong Jiang, Zongxin Yang, Zheqiao Cheng, Ruijie Quan, and Yi
  Yang.
\newblock {Insert Anything: Image Insertion via In-Context Editing in DiT}.
\newblock In \emph{Proceedings of the AAAI Conference on Artificial
  Intelligence (AAAI)}, pages 9097--9105, 2026{\natexlab{b}}.

\bibitem[Su et~al.(2024)Su, Ahmed, Lu, Pan, Bo, and Liu]{rope}
Jianlin Su, Murtadha Ahmed, Yu Lu, Shengfeng Pan, Wen Bo, and Yunfeng Liu.
\newblock {RoFormer: Enhanced Transformer with Rotary Position Embedding}.
\newblock \emph{Neurocomputing}, 568:\penalty0 127063, 2024.

\bibitem[Suhail et~al.(2022)Suhail, Esteves, Sigal, and Makadia]{gpnr}
Mohammed Suhail, Carlos Esteves, Leonid Sigal, and Ameesh Makadia.
\newblock {Generalizable Patch-Based Neural Rendering}.
\newblock In \emph{Proceedings of the European Conference on Computer Vision
  (ECCV)}, pages 156--174, 2022.

\bibitem[Tan et~al.(2025)Tan, Liu, Yang, Xue, and Wang]{ominicontrol}
Zhenxiong Tan, Songhua Liu, Xingyi Yang, Qiaochu Xue, and Xinchao Wang.
\newblock {OminiControl: Minimal and Universal Control for Diffusion
  Transformer}.
\newblock In \emph{Proceedings of the IEEE/CVF International Conference on
  Computer Vision (ICCV)}, pages 14940--14950, 2025.

\bibitem[Tang(2026)]{rgor}
Bangxun Tang.
\newblock {Beyond Lip Sync: Reference-Grounded Oral Refinement for Audio-Driven
  Portrait Animation}.
\newblock \emph{arXiv preprint arXiv:2609.38019}, 2026.

\bibitem[Tang et~al.(2026)Tang, Gao, Song, Fang, He, Yang, and Shou]{paiactor}
Bangxun Tang, Heyuan Gao, Yiren Song, Guian Fang, Zijian He, Jie Yang, and
  Mike~Zheng Shou.
\newblock {PAI-Actor: Cinematic Multi-Character Replacement in Dynamic Scenes}.
\newblock In \emph{Advances in Neural Information Processing Systems
  (NeurIPS)}, 2026.

\bibitem[Tang et~al.(2023)Tang, Jia, Wang, Phoo, and Hariharan]{dift}
Luming Tang, Menglin Jia, Qianqian Wang, Cheng~Perng Phoo, and Bharath
  Hariharan.
\newblock {Emergent Correspondence from Image Diffusion}.
\newblock In \emph{Advances in Neural Information Processing Systems
  (NeurIPS)}, pages 1363--1389, 2023.

\bibitem[Tang et~al.(2024)Tang, Ruiz, Chu, Li, Holynski, Jacobs, Hariharan,
  Pritch, Wadhwa, Aberman, and Rubinstein]{realfill}
Luming Tang, Nataniel Ruiz, Qinghao Chu, Yuanzhen Li, Aleksander Holynski,
  David~E. Jacobs, Bharath Hariharan, Yael Pritch, Neal Wadhwa, Kfir Aberman,
  and Michael Rubinstein.
\newblock {RealFill: Reference-Driven Generation for Authentic Image
  Completion}.
\newblock \emph{ACM Transactions on Graphics}, 43\penalty0 (4):\penalty0
  135:1--135:12, 2024.

\bibitem[{Team Seedance} et~al.(2026){Team Seedance}, Chen, Chen, Chen, Chen,
  Chen, Chen, Cheng, Cheng, Cheng, et~al.]{seedance2}
{Team Seedance}, De Chen, Liyang Chen, Xin Chen, Ying Chen, Zhuo Chen, Zhuowei
  Chen, Feng Cheng, Tianheng Cheng, Yufeng Cheng, et~al.
\newblock {Seedance 2.0: Advancing Video Generation for World Complexity}.
\newblock \emph{arXiv preprint arXiv:2604.14148}, 2026.

\bibitem[{Team Wan} et~al.(2025){Team Wan}, Wang, Ai, Wen, Mao, Xie, Chen, Yu,
  Zhao, Yang, et~al.]{wan}
{Team Wan}, Ang Wang, Baole Ai, Bin Wen, Chaojie Mao, Chen-Wei Xie, Di Chen,
  Feiwu Yu, Haiming Zhao, Jianxiao Yang, et~al.
\newblock {Wan: Open and Advanced Large-Scale Video Generative Models}.
\newblock \emph{arXiv preprint arXiv:2503.20314}, 2025.

\bibitem[Tewel et~al.(2025)Tewel, Gal, Samuel, Atzmon, Wolf, and
  Chechik]{addit}
Yoad Tewel, Rinon Gal, Dvir Samuel, Yuval Atzmon, Lior Wolf, and Gal Chechik.
\newblock {Add-it: Training-Free Object Insertion in Images With Pretrained
  Diffusion Models}.
\newblock In \emph{International Conference on Learning Representations
  (ICLR)}, 2025.

\bibitem[Tseng et~al.(2023)Tseng, Li, Kim, Alsisan, Huang, and
  Kopf]{poseguided}
Hung-Yu Tseng, Qinbo Li, Changil Kim, Suhib Alsisan, Jia-Bin Huang, and
  Johannes Kopf.
\newblock {Consistent View Synthesis With Pose-Guided Diffusion Models}.
\newblock In \emph{Proceedings of the IEEE/CVF Conference on Computer Vision
  and Pattern Recognition (CVPR)}, pages 16773--16783, 2023.

\bibitem[Tu et~al.(2025)Tu, Luo, Chen, Ji, Bai, and Zhao]{videoanydoor}
Yuanpeng Tu, Hao Luo, Xi Chen, Sihui Ji, Xiang Bai, and Hengshuang Zhao.
\newblock {VideoAnydoor: High-fidelity Video Object Insertion with Precise
  Motion Control}.
\newblock In \emph{ACM SIGGRAPH Conference Papers}, pages 1--11, 2025.

\bibitem[Unterthiner et~al.(2018)Unterthiner, van Steenkiste, Kurach, Marinier,
  Michalski, and Gelly]{fvd}
Thomas Unterthiner, Sjoerd van Steenkiste, Karol Kurach, Rapha{\"e}l Marinier,
  Marcin Michalski, and Sylvain Gelly.
\newblock {Towards Accurate Generative Models of Video: A New Metric \&
  Challenges}.
\newblock \emph{arXiv preprint arXiv:1812.01717}, 2018.

\bibitem[Wang et~al.(2026)Wang, Shi, Li, Bian, Liu, Lu, Wang, Wan, Gai, and
  Jia]{multishotmaster}
Qinghe Wang, Xiaoyu Shi, Baolu Li, Weikang Bian, Quande Liu, Huchuan Lu, Xintao
  Wang, Pengfei Wan, Kun Gai, and Xu Jia.
\newblock {MultiShotMaster: A Controllable Multi-Shot Video Generation
  Framework}.
\newblock In \emph{Proceedings of the IEEE/CVF Conference on Computer Vision
  and Pattern Recognition (CVPR)}, pages 16268--16278, 2026.

\bibitem[Wang et~al.(2004)Wang, Bovik, Sheikh, and Simoncelli]{ssim}
Zhou Wang, Alan~C. Bovik, Hamid~R. Sheikh, and Eero~P. Simoncelli.
\newblock {Image Quality Assessment: From Error Visibility to Structural
  Similarity}.
\newblock \emph{IEEE Transactions on Image Processing}, 13\penalty0
  (4):\penalty0 600--612, 2004.

\bibitem[Wei et~al.(2026)Wei, Liu, Ye, Wang, Wang, Wan, Gai, and
  Chen]{univideo}
Cong Wei, Quande Liu, Zixuan Ye, Qiulin Wang, Xintao Wang, Pengfei Wan, Kun
  Gai, and Wenhu Chen.
\newblock {UniVideo: Unified Understanding, Generation, and Editing for
  Videos}.
\newblock In \emph{International Conference on Learning Representations
  (ICLR)}, 2026.

\bibitem[Wu et~al.(2025{\natexlab{a}})Wu, Li, Zhou, Lin, Gao, Yan, Yin, Bai,
  Xu, Chen, et~al.]{qwenimage}
Chenfei Wu, Jiahao Li, Jingren Zhou, Junyang Lin, Kaiyuan Gao, Kun Yan,
  {Sheng-ming} Yin, Shuai Bai, Xiao Xu, Yilei Chen, et~al.
\newblock {Qwen-Image Technical Report}.
\newblock \emph{arXiv preprint arXiv:2508.02324}, 2025{\natexlab{a}}.

\bibitem[Wu et~al.(2026)Wu, Long, Huang, Sun, Chen, Guo, and
  Li]{thinkingonshots}
Chenyang Wu, Fuchen Long, Binyuan Huang, Xinlong Sun, Xi Chen, Chun-Le Guo, and
  Chongyi Li.
\newblock {Thinking on Shots: Consistent Multi-Shot Video Editing with Agentic
  Reasoning}.
\newblock \emph{arXiv preprint arXiv:2608.26809}, 2026.

\bibitem[Wu et~al.(2023)Wu, Ge, Wang, Lei, Gu, Shi, Hsu, Shan, Qie, and
  Shou]{tuneavideo}
Jay~Zhangjie Wu, Yixiao Ge, Xintao Wang, Stan~Weixian Lei, Yuchao Gu, Yufei
  Shi, Wynne Hsu, Ying Shan, Xiaohu Qie, and Mike~Zheng Shou.
\newblock {Tune-A-Video: One-Shot Tuning of Image Diffusion Models for
  Text-to-Video Generation}.
\newblock In \emph{Proceedings of the IEEE/CVF International Conference on
  Computer Vision (ICCV)}, pages 7623--7633, 2023.

\bibitem[Wu et~al.(2025{\natexlab{b}})Wu, Huang, Wu, Cheng, Ding, and He]{uno}
Shaojin Wu, Mengqi Huang, Wenxu Wu, Yufeng Cheng, Fei Ding, and Qian He.
\newblock {Less-to-More Generalization: Unlocking More Controllability by
  In-Context Generation}.
\newblock In \emph{Proceedings of the IEEE/CVF International Conference on
  Computer Vision (ICCV)}, pages 18682--18692, 2025{\natexlab{b}}.

\bibitem[Xia et~al.(2026)Xia, Cong, Yao, Wang, Ye, and Ma]{movi}
Qi Xia, Peishan Cong, Yichen Yao, Ziyi Wang, Yaoqin Ye, and Yuexin Ma.
\newblock {Controllable Video Object Insertion via Multi-View Priors}.
\newblock In \emph{Proceedings of the ACM International Conference on
  Multimedia (MM)}, pages 11933--11942, 2026.

\bibitem[Xiao et~al.(2026)Xiao, Yang, Zhang, Cai, Zhao, Guo, Wetzstein,
  Agrawala, Yuille, and Jiang]{captaincinema}
Junfei Xiao, Ceyuan Yang, Lvmin Zhang, Shengqu Cai, Yang Zhao, Yuwei Guo,
  Gordon Wetzstein, Maneesh Agrawala, Alan Yuille, and Lu Jiang.
\newblock {Captain Cinema: Towards Short Movie Generation}.
\newblock In \emph{International Conference on Learning Representations
  (ICLR)}, 2026.

\bibitem[Xu et~al.(2024)Xu, Nie, Liu, Liu, Kautz, Wang, and Vahdat]{camco}
Dejia Xu, Weili Nie, Chao Liu, Sifei Liu, Jan Kautz, Zhangyang Wang, and Arash
  Vahdat.
\newblock {CamCo: Camera-Controllable 3D-Consistent Image-to-Video Generation}.
\newblock \emph{arXiv preprint arXiv:2406.02509}, 2024.

\bibitem[Xu et~al.(2025)Xu, Huang, Pei, Hou, Li, Chen, Zhang, Feng, and
  Xie]{egoexogen}
Jilan Xu, Yifei Huang, Baoqi Pei, Junlin Hou, Qingqiu Li, Guo Chen, Yuejie
  Zhang, Rui Feng, and Weidi Xie.
\newblock {EgoExo-Gen: Ego-centric Video Prediction by Watching Exo-centric
  Videos}.
\newblock In \emph{International Conference on Learning Representations
  (ICLR)}, 2025.

\bibitem[Yang et~al.(2023)Yang, Gu, Zhang, Zhang, Chen, Sun, Chen, and
  Wen]{paintbyexample}
Binxin Yang, Shuyang Gu, Bo Zhang, Ting Zhang, Xuejin Chen, Xiaoyan Sun, Dong
  Chen, and Fang Wen.
\newblock {Paint by Example: Exemplar-Based Image Editing With Diffusion
  Models}.
\newblock In \emph{Proceedings of the IEEE/CVF Conference on Computer Vision
  and Pattern Recognition (CVPR)}, pages 18381--18391, 2023.

\bibitem[Yang et~al.(2026)Yang, Li, Cun, Wang, Li, Shan, and
  Zhang]{gencompositor}
Shuzhou Yang, Xiaoyu Li, Xiaodong Cun, Guangzhi Wang, Lingen Li, Ying Shan, and
  Jian Zhang.
\newblock {GenCompositor: Generative Video Compositing with Diffusion
  Transformer}.
\newblock In \emph{International Conference on Learning Representations
  (ICLR)}, 2026.

\bibitem[Yang et~al.(2025)Yang, Teng, Zheng, Ding, Huang, Xu, Yang, Hong,
  Zhang, Feng, et~al.]{cogvideox}
Zhuoyi Yang, Jiayan Teng, Wendi Zheng, Ming Ding, Shiyu Huang, Jiazheng Xu,
  Yuanming Yang, Wenyi Hong, Xiaohan Zhang, Guanyu Feng, et~al.
\newblock {CogVideoX: Text-to-Video Diffusion Models with An Expert
  Transformer}.
\newblock In \emph{International Conference on Learning Representations
  (ICLR)}, 2025.

\bibitem[Ye et~al.(2026)Ye, He, Liu, Wang, Wang, Wan, Zhang, Gai, Chen, and
  Luo]{unic}
Zixuan Ye, Xuanhua He, Quande Liu, Qiulin Wang, Xintao Wang, Pengfei Wan, Di
  Zhang, Kun Gai, Qifeng Chen, and Wenhan Luo.
\newblock {Unified In-Context Video Editing}.
\newblock In \emph{International Conference on Learning Representations
  (ICLR)}, 2026.

\bibitem[Yi et~al.(2018)Yi, Trulls, Ono, Lepetit, Salzmann, and Fua]{goodcorr}
Kwang~Moo Yi, Eduard Trulls, Yuki Ono, Vincent Lepetit, Mathieu Salzmann, and
  Pascal Fua.
\newblock {Learning to Find Good Correspondences}.
\newblock In \emph{Proceedings of the IEEE/CVF Conference on Computer Vision
  and Pattern Recognition (CVPR)}, pages 2666--2674, 2018.

\bibitem[Yu et~al.(2025)Yu, Hu, Xing, and Shan]{trajectorycrafter}
Mark Yu, Wenbo Hu, Jinbo Xing, and Ying Shan.
\newblock {TrajectoryCrafter: Redirecting Camera Trajectory for Monocular
  Videos via Diffusion Models}.
\newblock In \emph{Proceedings of the IEEE/CVF International Conference on
  Computer Vision (ICCV)}, pages 100--111, 2025.

\bibitem[Zhang et~al.(2018)Zhang, Isola, Efros, Shechtman, and Wang]{lpips}
Richard Zhang, Phillip Isola, Alexei~A. Efros, Eli Shechtman, and Oliver Wang.
\newblock {The Unreasonable Effectiveness of Deep Features as a Perceptual
  Metric}.
\newblock In \emph{Proceedings of the IEEE/CVF Conference on Computer Vision
  and Pattern Recognition (CVPR)}, pages 586--595, 2018.

\bibitem[Zhang et~al.(2025)Zhang, Yuan, Song, Wang, and Liu]{easycontrol}
Yuxuan Zhang, Yirui Yuan, Yiren Song, Haofan Wang, and Jiaming Liu.
\newblock {EasyControl: Adding Efficient and Flexible Control for Diffusion
  Transformer}.
\newblock In \emph{Proceedings of the IEEE/CVF International Conference on
  Computer Vision (ICCV)}, pages 19513--19524, 2025.

\bibitem[Zhou et~al.(2021)Zhou, Barnes, Shechtman, and Amirghodsi]{transfill}
Yuqian Zhou, Connelly Barnes, Eli Shechtman, and Sohrab Amirghodsi.
\newblock {TransFill: Reference-Guided Image Inpainting by Merging Multiple
  Color and Spatial Transformations}.
\newblock In \emph{Proceedings of the IEEE/CVF Conference on Computer Vision
  and Pattern Recognition (CVPR)}, pages 2266--2276, 2021.

\bibitem[Zhuang et~al.(2025)Zhuang, Huang, Yang, Zhang, Wang, Fu, Sun, Zha, Li,
  and Wang]{getinvideo}
Shaobin Zhuang, Zhipeng Huang, Binxin Yang, Ying Zhang, Fangyikang Wang,
  Canmiao Fu, Chong Sun, Zheng-Jun Zha, Chen Li, and Yali Wang.
\newblock {Get In Video: Add Anything You Want to the Video}.
\newblock \emph{arXiv preprint arXiv:2503.06268}, 2025.

\bibitem[Zi et~al.(2025)Zi, Peng, Qi, Wang, Zhao, Xiao, and
  Wong]{minimaxremover}
Bojia Zi, Weixuan Peng, Xianbiao Qi, Jianan Wang, Shihao Zhao, Rong Xiao, and
  Kam-Fai Wong.
\newblock {MiniMax-Remover: Taming Bad Noise Helps Video Object Removal}.
\newblock In \emph{Advances in Neural Information Processing Systems
  (NeurIPS)}, 2025.

\end{thebibliography}
